\documentclass[twoside,11pt]{article}
\usepackage{jmlr2e}

\usepackage{sidecap}
\usepackage[printonlyused]{acronym}
\usepackage[export]{adjustbox} %
\usepackage{algorithm}
\usepackage{algorithmicx} %
\usepackage{algpseudocode} %
\usepackage[table]{xcolor} %
\algrenewcommand\algorithmicindent{0.8em}%
\usepackage{amssymb}
\usepackage{amsmath}
\usepackage{array}
\usepackage{bbm}
\usepackage{booktabs}
\usepackage{caption}
\usepackage{float}
\usepackage[table]{xcolor}
\usepackage{graphicx}
\usepackage{hyperref}
\usepackage{lipsum}
\usepackage{listings}
\usepackage{microtype}
\usepackage{multicol}
\usepackage{multirow}
\usepackage{natbib}
\usepackage{nicefrac}
\usepackage{relsize}
\usepackage{siunitx}
\usepackage{subcaption}
\usepackage{tikz}
\usetikzlibrary{shapes, arrows, positioning, fit, backgrounds}
\usetikzlibrary{calc}
\usetikzlibrary{decorations.pathmorphing}
\tikzstyle{block} = [rectangle, draw, fill=blue!20,
    text centered, rounded corners, minimum height=1em, node distance=1.5cm]
\tikzstyle{block2} = [rectangle, draw, fill=magenta!60,
    text centered, rounded corners, minimum height=1em, node distance=1.5cm]
\tikzstyle{line} = [draw, -latex']
\tikzstyle{cloud} = [draw, rectangle,fill=red!20, node distance=1.5cm and 1cm,
    minimum height=1em]
\usepackage{verbatim}
\usepackage{xspace}
\usepackage{wrapfig}

\definecolor{mydarkblue}{rgb}{0,0.08,0.45}
\hypersetup{ %
    pdftitle={},
    pdfauthor={},
    pdfsubject={},
    pdfkeywords={},
    pdfborder=0 0 0,
    pdfpagemode=UseNone,
    colorlinks=true,
    linkcolor=mydarkblue,
    citecolor=mydarkblue,
    filecolor=mydarkblue,
    urlcolor=mydarkblue,
    pdfview=FitH}

\usepackage{array,multirow}
\usepackage{cprotect}
\usepackage{csquotes}

\usetikzlibrary{tikzmark}
\usetikzlibrary{calc}

\errorcontextlines\maxdimen

\newcommand{\ALGtikzmarkcolor}{black}%
\newcommand{\ALGtikzmarkextraindent}{2pt}%
\newcommand{\ALGtikzmarkverticaloffsetstart}{-.5ex}%
\newcommand{\ALGtikzmarkverticaloffsetend}{-.5ex}%
\makeatletter
\newcounter{ALG@tikzmark@tempcnta}

\newcommand\ALG@tikzmark@start{%
    \global\let\ALG@tikzmark@last\ALG@tikzmark@starttext%
    \expandafter\edef\csname ALG@tikzmark@\theALG@nested\endcsname{\theALG@tikzmark@tempcnta}%
    \tikzmark{ALG@tikzmark@start@\csname ALG@tikzmark@\theALG@nested\endcsname}%
    \addtocounter{ALG@tikzmark@tempcnta}{1}%
}

\def\ALG@tikzmark@starttext{start}
\newcommand\ALG@tikzmark@end{%
    \ifx\ALG@tikzmark@last\ALG@tikzmark@starttext
    \else
        \tikzmark{ALG@tikzmark@end@\csname ALG@tikzmark@\theALG@nested\endcsname}%
        \tikz[overlay,remember picture] \draw[\ALGtikzmarkcolor] let \p{S}=($(pic cs:ALG@tikzmark@start@\csname ALG@tikzmark@\theALG@nested\endcsname)+(\ALGtikzmarkextraindent,\ALGtikzmarkverticaloffsetstart)$), \p{E}=($(pic cs:ALG@tikzmark@end@\csname ALG@tikzmark@\theALG@nested\endcsname)+(\ALGtikzmarkextraindent,\ALGtikzmarkverticaloffsetend)$) in (\x{S},\y{S})--(\x{S},\y{E});%
    \fi
    \gdef\ALG@tikzmark@last{end}%
}

\apptocmd{\ALG@beginblock}{\ALG@tikzmark@start}{}{\errmessage{failed to patch}}
\pretocmd{\ALG@endblock}{\ALG@tikzmark@end}{}{\errmessage{failed to patch}}
\makeatother

\firstpageno{1}

\newcommand{\changes}[1]{#1} %
\newcommand{\newchanges}[1]{#1} %
\newcommand{\notice}[1]{{\color{blue}{#1}}~}

\newcommand{\refSection}[1]{Sec. \ref{#1}}
\newcommand{\refFigure}[1]{Fig.~\ref{#1}}

\newcommand{\refAlgorithm}[1]{Alg.~\ref{#1}}
\newcommand{\appendixx}[0]{Appendix\xspace}

\newcommand{\normalDistribution}[2]{\mathcal{N}(#1,#2)}
\newcommand{\expectation}{\mathbb{E}}

\newcommand{\videoLink}{https://sites.google.com/view/virl1}

\newcommand{\distanceMetricText}{distance metric\xspace}

\newcommand{\demonstrationText}{demonstration\xspace}
\newcommand{\methodName}{VIRL\xspace}
\newcommand{\humanoidThreeD}{humanoid3d\xspace}
\newcommand{\humanoidTwoD}{humanoid2d\xspace}

\newcommand{\stateSpace}{S}
\newcommand{\actionSpace}{A}

\newcommand{\myState}{\sstate}
\newcommand{\sstate}{s}
\newcommand{\observation}{\textbf{o}}
\newcommand{\observationSeq}{O}
\newcommand{\action}{a}
\newcommand{\reward}{r}
\newcommand{\ttime}{t}

\newcommand{\latentVariable}{\textbf{z}}

\newcommand{\discountFactor}{\gamma}

\newcommand{\agent}{agent\xspace}

\newcommand{\policySymbol}{\pi}

\newcommand{\expert}{expert\xspace}
\newcommand{\encoderSymbol}{\phi}
\newcommand{\decoderSymbol}{\psi}

\newcommand{\policy}[1]{\policySymbol(#1)}

\newcommand{\distance}[1]{d(#1)}

\newcommand{\modelParametersPolicy}[0]{\theta_{\policySymbol}}

\newcommand{\tripletMargin}[0]{\rho}
\newcommand{\trajectory}[0]{\tau}

\usepackage{lastpage}
\jmlrheading{24}{2023}{1-\pageref{LastPage}}{10/21; Revised
1/23}{3/23}{21-1174}{Glen Berseth, Florian Golemo, Christopher Pal}
\ShortHeadings{Towards Learning to Imitate from Video}{Berseth et al.}

\begin{document}

\title{Towards Learning to Imitate \\ from a Single Video Demonstration}

\author{\name Glen Berseth 
\addr Université de Montréal, Mila Quebec AI Institute, and Canada CIFAR AI Chair
\email glen.berseth@mila.quebec \\[-15pt]
      \AND
      \name Florian Golemo
      \addr Mila Quebec AI Institute
      \email fgolemo@gmail.com \\[-15pt]
    \AND \name Christopher Pal
    \addr Polytechnique Montréal, Mila Quebec AI Institute, ServiceNow Research, and Canada CIFAR AI Chair
    \email christopher.pal@servicenow.com
       }
\editor{George Konidaris}

\maketitle

\medskip

  \acrodef{MDP}{Markov decision process}

\acrodef{DDPG}{deep deterministic policy gradient}
\acrodef{ReLU}{rectified linear unit}
\acrodef{PPO}{proximal policy optimization}
\acrodef{RL}{reinforcement learning}
\acrodef{GAN}{generative adversarial network}
\acrodef{RNN}{recurrent neural network}
\acrodef{LSTM}{long short-term memory}
\acrodef{BC}{behavioural cloning}
\acrodef{IRL}{inverse reinforcement learning}
\acrodef{GAIL}{generative adversarial imitation learning}
\acrodef{GAIfO}{generative adversarial imitation from observation}
\acrodef{COM}{centre of mass}
\acrodef{t-SNE}{t-distributed stochastic neighbor embedding}
\acrodef{DoF}{degrees of freedom}
\acrodef{RSAE}{recurrent sequence autoencoder}
\acrodef{TRPO}{trust-region policy optimization}
\acrodef{GPU}{graphics processing unit}
\acrodef{TCN}{time-contrastive network}
\acrodef{VAE}{variational autoencoder}
\acrodef{AE}{autoencoder}
\acrodef{RSI}{reference state initialization}
\acrodef{EESP}{early episode sequence priority}
\acrodef{LfD}{learning from demonstration}
\acrodef{CACLA}{continuous actor-critic learning-automaton}
\begin{abstract}
Agents that can learn to imitate behaviours observed in video -- \emph{without having direct access to internal state or action information of the observed agent} -- are more suitable for learning in the natural world. However, formulating a reinforcement learning (RL) agent that facilitates this goal remains a significant challenge.
We approach this challenge using contrastive training to learn a reward function by comparing an agent's behaviour with a single demonstration.
We use a Siamese recurrent neural network architecture to learn rewards in space and time between motion clips while training an RL policy to minimize this distance.
Through experimentation, we also find that the inclusion of multi-task data and additional image encoding losses improve the temporal consistency of the learned rewards and, as a result, significantly improve policy learning.
We demonstrate our approach on simulated humanoid, dog, and raptor agents in 2D and quadruped and humanoid agents in 3D.  We show that our method outperforms current state-of-the-art techniques and can learn to imitate behaviours from a single video demonstration.
\end{abstract}


\section{Introduction}
\label{sec:Intro}

\begin{wrapfigure}{r}{0.4\linewidth}
\vspace{-1.0cm}
    \centering
    \includegraphics[width=0.9\linewidth]{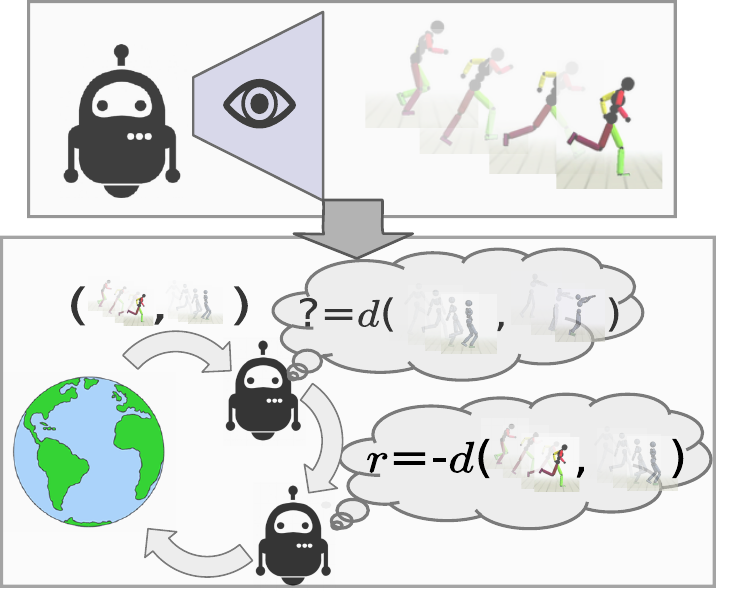}
    \caption{The agent visually observes behaviour and then collects experience to train a distance function $d()$, used to learn how to reproduce that behaviour with RL.}
    \label{fig:virl_teaser}
\vspace{-0.5cm}
\end{wrapfigure}
Imitation learning gives an agent the ability to reproduce the behaviours and skills of other agents through demonstrations~\citep{blakemore2001perception}. These demonstrations act as a type of explicit communication, informing the agent of the desired behaviour.
However, real-world agents such as robots do not generally have access to the type of information needed by many imitation learning methods, such as internal state information or the executed actions of a demonstration. 
We also want a solution that can learn to imitate even if an observed agent has a different appearance or dynamics in the demonstration compared to the agent that is tasked with learning from the demonstration. In the same way that human children can learn to imitate adults by observing them, we want more versatile agents who can learn to imitate desired behaviours, given only a few examples.
This type of visual-imitation is depicted in~\autoref{fig:virl_teaser} with an agent that learns and minimizes the distance between visual demonstrations.
\changes{If we can construct agents that can learn to imitate from this kind of easy-to-obtain but noisy data, they would be much more flexible for learning via imitation in the real world.}

While imitating a behaviour from observation is natural to many agents in the real world, it poses many learning challenges.
Behaviour cloning (BC) methods often require expert action data making them impossible to use in more natural problem settings where only image observations are available. 
Also, in the real world, the demonstration agent often has different dynamics, meaning that an exact copy of the demonstration is impossible, and the learning agent must do its best to replicate the observed behaviour under \textit{its own} dynamics.
How can we train an agent to reliably imitate another agent with potentially different dynamics, given only image data from the demonstration agent?
Learning a well-formed and smooth distance between observed behaviours can be used as the reward for a reinforcement learning agent.
However, given the partially observed nature of the demonstration data, learning a reasonable distance function is challenging. To compensate for limited and noisy data, a method that allows us to incorporate additional offline data, potentially from other behaviours/tasks, will increase data efficiency while training a model that understands the comparative landscape of a larger behaviour space.

\begin{figure*}[tb]
\vspace{-.25cm}
\begin{subfigure}{0.6\textwidth}
\centering
\includegraphics[width=0.8\linewidth]{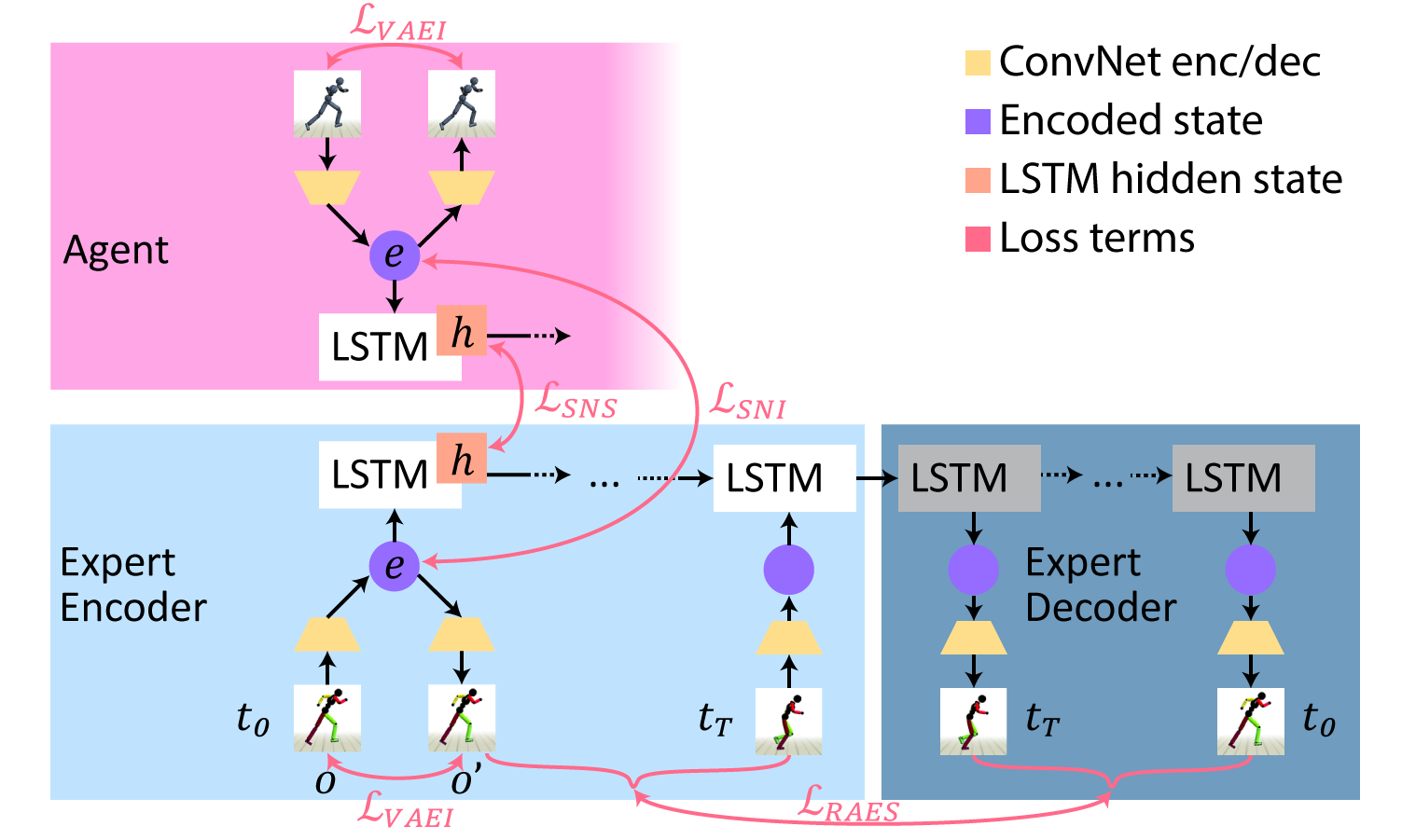} 
\caption{Losses for training the encoders/decoders.}
\label{fig:overview-losses}
\end{subfigure}
\begin{subfigure}{0.35\textwidth}
\centering
\includegraphics[width=0.945\linewidth]{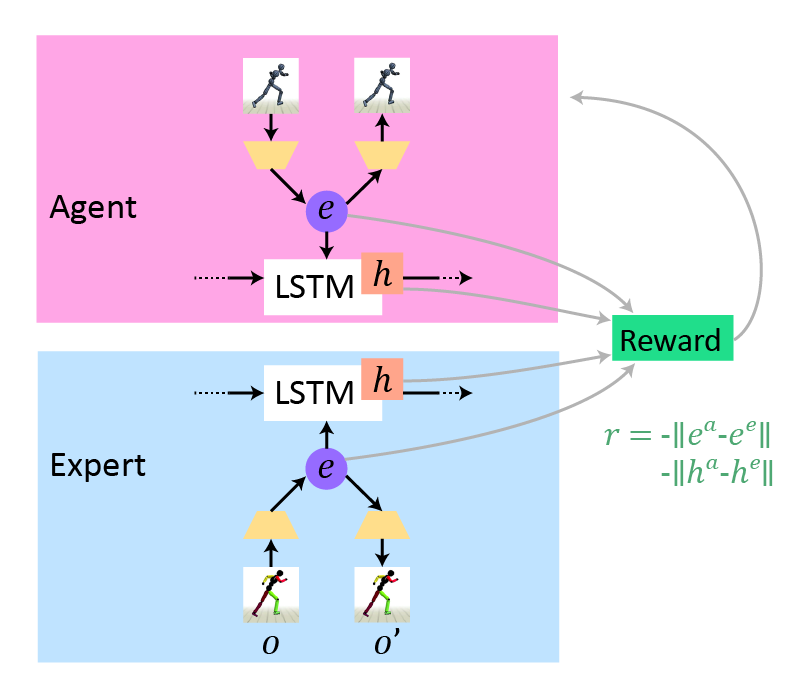}
\caption{Reward generation for the agent.}
\label{fig:overview-rewards}
\end{subfigure}
\caption{\methodName learns a distance function (\ref{fig:overview-losses}) and then uses that distance function as a reward function for RL (\ref{fig:overview-rewards}). At the current timestep, observations ($\observation$) of the reference motion and the agent are encoded ($\mathbf{e}$) and fed into \ac{LSTM}s (leading to hidden states $\mathbf{h}$). Fig. \ref{fig:overview-losses} shows how the reward model is trained using both Siamese and \ac{AE} losses. There are \ac{VAE} reconstruction losses on static images ($\mathcal{L}_{VAEI}$),  sequence-to-sequence AE losses  ($\mathcal{L}_{RAES}$), one for the reference and one for the agent (which we do not show in pink to simplify the figure). There is a Siamese loss between encoded images ($\mathcal{L}_{SNI}$) and a Siamese loss that is computed between encoded states over time ($\mathcal{L}_{SNS}$).  \refFigure{fig:overview-rewards} shows how the reward is calculated \textbf{at every timestep} using the combination of the learned encodings. 
}
\label{fig:overview}
\vspace{-0.25cm}
\end{figure*}

To realize the data-efficient and task-independent method described above, we train a recurrent Siamese comparator to capture the partial information from each image and use this comparator to compute distances used for rewards for an RL agent.
We train this comparator model with off-policy data to learn distances between sequences of images (videos). This offline training makes it possible to pretrain and include data from additional behaviours/tasks to increase model robustness.
Auto-encoding losses are added, shown in~\refFigure{fig:overview-losses}, at different levels of granularity to increase the smoothness of the learned distance landscape.
Our model learns two latent distance predictors in parallel. These two latent distance predictors are shown in~\refFigure{fig:overview-rewards} and allow us to compute distances between individual images and between sequences.
The \textit{image-to-image} latent distance rewards the agent for precisely matching the example behaviour. In contrast, the learned \textit{sequence-to-sequence} latent representation provides additional reward when the agent is not just matching the desired behaviour exactly but also when the agent is replicating portions of the observed demonstration, for example, if the agent currently has different timing than the demonstration.
Our results show that the robustness and smoothness gained from the combination of losses improve training efficiency and final policy quality.

Our contribution consists of a new visual-imitation learning method for RL based on visual comparisons and the specific architectures and training procedures discussed in more detail throughout the paper.
We showcase that our approach enables agents to learn a large variety of challenging behaviours that include walking, running and jumping. 
We perform experiments for multiple simulated robots in both 2D and 3D, including \changes{simulations for training quadruped robots} and a humanoid with $38$ \ac{DoF}. For many of these imitation tasks, our method ``visual-imitation with reinforcement learning" (\methodName) is able to imitate the given skill using only a single observed demonstration.

\section{Related Work}

We group the most relevant prior work based on the type and quantity of data needed to perform imitation learning. 
The first group consists of GAIL~\citep{NIPS2016_6391} and related methods, which require access to expert policies, states, actions, and require large quantities of expert data. 
In the second group, the need for expert action data is relaxed in methods like \ac{GAIfO}~\citep{torabi2018generative}, but still, requires an expert policy to repeatedly generate data. 
The third group avoids the need for ground truth states favouring images that are easier to obtain~\citep{brown2019extrapolating,brown2020better}. 
These methods still require many examples of data from a policy trained on an agent in the same simulation with the same dynamics.
Lastly, in a fourth group, the need for multiple demonstrations and matching dynamics is relaxed as in methods such as \ac{TCN}~\citep{sermanet2018time} and ours.

\paragraph{Imitation learning.} Methods
 such as \ac{GAIL}~\citep{NIPS2016_6391},  use the \ac{GAN} ~\citep{NIPS2014_5423} framework and applies it in the context of learning an RL policy. 
In GAIL, the \ac{GAN}'s discriminator is trained with positive examples from expert trajectories and negative examples from the current policy.
However, using a discriminator is only one possible way of measuring the probability of that agent's behaviour matching the expert~\citep{Abbeel:2004:ALV:1015330.1015430,ARGALL2009469, DBLP:journals/corr/FinnYFAL16, brown2019extrapolating, NEURIPS2021_fd00d347}.
Given a vector of features, distance-based imitation learning aims to find an optimal transformation that computes a more meaningful distance between expert demonstrations and agent trajectories.
Previous work has explored the area of state-based distance functions, but most rely on the availability of an \expert policy to continuously sample data~\citep{NIPS2016_6391,DBLP:journals/corr/MerelTTSLWWH17}. 
In the section hereafter, we demonstrate how \methodName learns a more stable distance-based reward over sequences of images (as opposed to states) and without access to actions or \expert policies.

\paragraph{Imitation without action data.}
For learning from demonstrations (LfD) problems, the goal is to replicate the behaviour of an expert $\policySymbol_{E}$.  
\ac{GAIfO}~\citep{torabi2018generative} has been proposed as an extension of GAIL that does not require data on the \expert actions. However, 
\ac{GAIfO} and other recent works in this area require access to an expert policy for sampling additional states during training~\citep{DBLP:conf/icml/0002VBB19,NIPS2019_8317}. 
By comparison, our method can work with a single fixed demonstration example with different dynamics.
Other recent work uses \ac{BC} to learn an inverse dynamics model to estimate the actions used via maximum-likelihood estimation~\citep{Torabi2018}.
Still, \ac{BC} often needs many expert examples and tends to suffer from state distribution mismatch issues between the \textit{expert} policy and \textit{student}~\citep{dagger}.

Additional works learn implicit models of distance that require large amounts of demonstration data and none of these explicitly learn a sequential model considering the \demonstrationText timing~\citep{DBLP:journals/corr/abs-1802-01557,DBLP:journals/corr/abs-1709-04905,sermanet2018time,DBLP:journals/corr/MerelTTSLWWH17,edwards2019imitating,NIPS2019_8528}.
The work in~\citet{NIPS2017_7116,NIPS2017_6971,peng2018variational,2021-TOG-AMP} includes a more robust \ac{GAIL} framework with a new model to encode motions for few-shot imitation. However, they need access to an expert policy to sample additional data.
In this work, we train recurrent siamese networks~\citep{chopra2005learning} to learn more meaningful distances between videos. Other work uses state-only demonstration to out-perform the demonstration data but requires many demonstrations and ranking information to be successful~\citep{brown2019extrapolating,brown2020better}.
We show results on more complex 3D tasks and additionally model distance in time, i.e. due to the embedding of the entire sequence, our model can compute meaningful distances between \agent and \demonstrationText even if they deviate in time.

\paragraph{Imitation from images.}
Some works like~\citet{DBLP:journals/corr/abs-1709-04905,sermanet2018time,DBLP:journals/corr/LiuGAL17,2018arXiv180800928D}, use image-based inputs instead of states but require on the order of hundreds of demonstrations. 
Further, these models only address spatial alignment, matching joint positions/orientations to a single state, and can not implicitly provide an additional signal related to temporal ordering between expert demonstration and agent motion like our recurrent sequence model does.
Other works that imitate image-based information do so only between goal states~\citep{
DBLP:journals/corr/abs-1804-08606}.

\paragraph{Imitation from few images with different dynamics.}
Comparative methods like \ac{TCN} use metric learning to embed \emph{simultaneous viewpoints} of the same object~\citep{sermanet2018time}. They use \ac{TCN} embeddings as \emph{features in the system state} which are provided to PILQR~\citep{chebotar2017combining} reinforcement learning algorithm, which combines model-based learning, linear time-varying dynamics and model-free corrections. In contrast, our Siamese network-based approach \emph{learns the reward} for an arbitrary subsequent RL algorithm. Our method does not rely on multiple views, and we use a \ac{RNN}-based autoencoding approach to regularize the distance computations used for generated rewards. These choices allow \methodName to achieve performance gains over \ac{TCN}, as shown in Section~\ref{sec:results}.

\section{Preliminaries}
\label{sec:framework}

In this section, we provide a very brief review of the fundamental background used by our method. \ac{RL} is formulated within the framework of a \ac{MDP} where at every time step $\ttime$, the world (including the \agent) exists in a state $ \myState_{\ttime} 
\in \stateSpace $, where the agent is able to perform actions $ \action_{\ttime} \in 
\actionSpace $. The action to take is determined according to a policy $ \policy{\action_{\ttime}|\myState_{\ttime}}$ which
results in a new state $ \myState_{\ttime+1} \in \stateSpace $  and reward $\reward_{t} = R(\myState_t, \action_t, \myState_{\ttime+1})$ according to the transition 
probability function $ P(\myState_{\ttime+1} | \myState_{\ttime}, \action_{\ttime}) $. 
The policy is optimized to maximize the future discounted reward
$
  \expectation_{\reward_{0}, ..., \reward_T} \left[ \sum_{\ttime=0}^{T} \gamma^\ttime \reward_{\ttime} \right]$,
\noindent where $ T $ is the max time horizon, and $ \discountFactor $ is the 
discount factor.
The formulation above generalizes to continuous states and actions, which is the situation for the agents we consider in our work.

\paragraph{Imitation Learning.}
Imitation learning is typically cast as the process of training a new policy to reproduce expert policy behaviour.
Behaviour cloning is a fundamental method for imitation learning.
Given an expert policy $\policySymbol_{E}$ possibly represented as a collection of trajectories $\trajectory = \langle (\myState_{0}, \action_{0}), \ldots, (\myState_{T}, \action_{T}) \rangle$ a new policy $\policySymbol$ can be learned to match this trajectory using supervised learning and maximizing the expectation
$\expectation_{\policySymbol_{E}}\left[\sum_{\ttime = 0}^{T} \log \policySymbol(\action_{\ttime}| \myState_{\ttime}, \modelParametersPolicy )\right]$.
While this simple method can work well, it often suffers from distribution mismatch issues leading to compounding errors as the learned policy deviates from the expert's behaviour~\citep{ross2011reduction}.        
Inverse reinforcement learning avoids this issue by extracting a reward function from observed optimal behaviour \citep{ng2000algorithms}.
In our approach, we learn a distance function that allows an agent to compare an observed behaviour to its current behaviour to define its reward $r_t$ at a given time step. 
Our method only requires a single reference activity, but the comparison network can be trained across a collection of different behaviours. 
Further, we do not assume the example data to be optimal.
See Appendix~\ref{sec:IRL} for further details of the connections of our work to inverse reinforcement learning.

\paragraph{Variational Auto-encoders}
\ac{VAE}s 
are a popular approach for learning lower-dimensional representations of a distribution~\citep{kingma2013auto}.
\newchanges{A \ac{VAE} consists of two parts, an encoder $q(\textbf{z} | \textbf{x}, \encoderSymbol)$, with parameters $\encoderSymbol$
and a decoder $p(\hat{\textbf{x}} | \latentVariable, \decoderSymbol)$ with parameters $\decoderSymbol$. The encoder maps inputs $\textbf{x}$ to a latent encoding $\latentVariable$, and, in turn, the decoder transforms $\latentVariable$ back to a reconstruction $\hat{\textbf{x}}$.
The model parameters for both $\encoderSymbol$ and $\decoderSymbol$ are trained jointly to maximize 
\begin{equation}
\label{eq:vae-loss}
    \mathcal{L}_{VAE}(\myState, \encoderSymbol, \decoderSymbol)  = -%
    \newchanges{D_{KL}(q(\latentVariable | \textbf{x}, \encoderSymbol)| p(\latentVariable)) + 
     \expectation_{q(\latentVariable | \textbf{x}, \encoderSymbol)}[\log p(\textbf{x} | \latentVariable, \decoderSymbol)]},
\end{equation}
where $D_{KL}$ is the Kullback-Leibler divergence and $p(\latentVariable)$ is a prior distribution over the latent space. %
The encoder (inference model) takes the form of a multivariate diagonal covariance distribution $q(\textbf{z} | \textbf{x}, \encoderSymbol) = \normalDistribution{\mu(\cdot| \textbf{x}, \encoderSymbol)}{\sigma^{2}(\cdot| \textbf{x}, \encoderSymbol)}$, where the mean $\mu( \cdot | \textbf{x}, \encoderSymbol)$ and variance $\sigma^{2}(\cdot| \textbf{x}, \encoderSymbol)$ are typically given by a deep neural network.

VAEs have been extended to work on sequences of images with the inclusion of an RNN, e.g. in \cite{chung2015recurrent} and the notation in the following section follows the one from that work. 
Sequence-to-sequence models can be used to learn the conditional probability of one sequence given another $p(y_{0}, \ldots, y_{T'}| x_{0}, \ldots, x_{T})$, where $\textbf{x} = x_{0}, \ldots, x_{T}$ and $ \textbf{y} = y_{0}, \ldots, y_{T'}$ are sequences.
Here, we will use extensions of encoder-decoder recurrent neural networks which learn a latent representation $\textbf{h}$ that compresses the information. 
For \methodName we reuse the VAE encoder mean from~\autoref{eq:vae-loss} to encode individual images $e_t \leftarrow \mu(x_{t}, \encoderSymbol) $ for which we learn the sequence encoder $f$, denoted as  $\mathbf{h} \leftarrow f(e_{0}, \ldots, e_{T}, \omega)$ with parameters $\omega$.
Conversely, a sequence decoder $g$, conditioned on $\textbf{h}$, reconstructs the original input sequence $x_{0}, \ldots, x_T$. First by producing a decoded sequence of the correct length $y_0, \cdots, y_t \leftarrow g(\textbf{h}, \rho) $ with parameters $\rho$ and second, using the decoder from~\autoref{eq:vae-loss} by $\hat{\textbf{x}}_t \leftarrow p(\textbf{y}_t, \decoderSymbol)$.
The loss for decoding the original sequence $\textbf{X} = \{x_0, \ldots, x_t\}$ can then be written as the $L_2$ distance}
\begin{equation}
\label{eq:sequence-autoencoder}
    \newchanges{\mathcal{L}_{RAES}(\textbf{X}, \encoderSymbol, \decoderSymbol, \omega, \rho) = || \textbf{X} -  
    \{p(\textbf{y}_0, \decoderSymbol), \ldots, p(\textbf{y}_t, \decoderSymbol)\}||^2}
\end{equation}
This method works for learning compressed representations for transfer learning~\citep{zhu2016deep} and 3D shape retrieval~\citep{zhuang2015supervised}. In our case, this type of autoencoding can help regularize our model by forcing the encoding to contain all the information needed to reconstruct the trajectory.

\section{Visual Imitation with Reinforcement Learning}
\label{sec:viz-imitation-method}

\changes{In this work, we create a new method for performing imitation from only visual data (no actions) and use reinforcement learning to fill in the missing \newchanges{action} data by training the agent to find the actions that result in matching the original distribution of observations.}

\paragraph{The Sequence Encoder/Decoder Networks}
\label{subsection:method-encdec-networks}
Figure \ref{fig:overview-losses} shows an outline of the network design. 
\newchanges{There are 2 LSTMs, one for sequence encoding and one for sequence decoding, as well as one encoder CNN, and one decoder CNN, all of which are shared across the agent and expert, similar to a Siamese network.
A single convolutional network $e^e_{\ttime} \leftarrow \mu(\textbf{\observation}_{\ttime}; \phi)$ is used to transform observations (images) at every timestep $\ttime$ of the \demonstrationText from the expert $\textbf{\observation}^{e}_{\ttime}$ or the agent $\textbf{\observation}^{a}_{\ttime}$ to the corresponding encoding vector $e_{\ttime}$.
After the observations are passed through the image encoder, the result is an encoded sequence $\langle e^{e}_{0}, \ldots, e^{e}_{\ttime} \rangle$ (for the expert in this example), this sequence is fed into the  $\texttt{LSTM}$ sequence encoder until a final encoding is produced $\textbf{h}^e_{\ttime} \leftarrow f(e^e_{0}, \ldots, e^e_{T}; \omega)$.
This same process is performed over the agent data $\textbf{\observation}^{a}$ producing $\textbf{h}^a_{\ttime} \leftarrow f(e^a_{0}, \ldots, e^a_{T}; \omega)$. These sequence encodings $\textbf{h}^e_{\ttime}$ and $\textbf{h}^a_{\ttime}$ are fed into the sequence decoder $g(\textbf{h}_{\ttime}; \rho)$ separately, which generates a series of decoded latent representations $ \langle \hat{y}^{e}_{0}, \ldots, \hat{y}^{e}_{\ttime} \rangle$ which are then decoded back to images with a deconvolutional network $\hat{o} \leftarrow p(\hat{y}^{e}_{\ttime}; \psi)$. The same process is applied to both the \agent and \expert using the same image encoder $\phi$ and decoder $\psi$ and sequence encoder $\omega$, and  decoder $\rho$.}

\paragraph{Loss Terms}
\label{subsection:method-losses}
The siamese-loss between a fully encoded demonstration sequence $h^e_{\ttime}$ and a sequence of the \agent $h^a_{\ttime}$ forces not just individual frames but the representation of entire sequences to match if they are from the same policy. This siamese network sequence loss $\mathcal{L}_{SNS}$ is defined in Eq.\ref{eq:siamese-loss}. A frame-by-frame siamese-loss between $e^e_{\ttime}$ of the \demonstrationText and $e^a_{\ttime}$ of the \agent encourages individual frames to have similar encodings as well. This siamese network image loss $\mathcal{L}_{SNI}$ uses Eq.\ref{eq:siamese-loss} as well but is trained over pairs of images. We define The Siamese network loss (both for images and sequences) as:
\begin{equation}
\begin{split}
    \label{eq:siamese-loss}
    \mathcal{L}_{SNX}(\observation_{i}, \observation_{p}, \changes{\observation_{n}}, y; \phi) 
     = & y * ||f(\observation_{i}; \phi) - f(\observation_{p}; \phi)||^2 + \\
    &  (1 - y) *  
    (\max( \tripletMargin - (||f(\observation_{i}; \phi) - f(\observation_{n}; \phi)||^2), 0)),
\end{split}
 \end{equation}
where $y \in [0,1]$ is the indicator for negative/positive samples. When
$y = 1$, the pair is positive, and the distance between current observation  $\observation_{i}$ to positive sample $\observation_{p}$ should be minimal. When $y = 0$, the pair is negative, and the distance between $\observation_{i}$ and negative example $\observation_{n}$ should be increased. \changes{In \autoref{sec:data-augmentation} the details of how the positive and negative examples are constructed is outlined.} We compute this loss over batches of data that are half positive and half negative pairs.
The margin $\tripletMargin$ \changes{is set to $1$ and} is used as an attractor or anchor to pull the negative example output away from $\observation_{i}$ and push values towards a $[0,1]$ range. $f(\cdot)$ computes the output from the underlying network (i.e. $\texttt{Conv}$ or $\texttt{LSTM}$).
The data used to train the Siamese network is a combination of observation trajectories $\textbf{\observationSeq} = \langle\observation_{0}, \ldots, \observation_{T}\rangle$ generated from simulating the \agent in the environment and the \demonstrationText.
For our recurrent model the observations $\textbf{\observationSeq}_{p}, \textbf{\observationSeq}_{n}, \textbf{\observationSeq}_{i}$ are sequences. 
We additionally train the encoding of a single observation of either \agent or \expert at a given timestep using the \ac{VAE} loss $\mathcal{L}_{VAE}$ from Eq.\ref{eq:vae-loss}. Lastly, the entire sequence of observations of both the \agent and \expert is encoded and then decoded back separately, as shown in \refFigure{fig:siamese-models}, and the $\texttt{LSTM}s$ are trained with the loss $\mathcal{L}_{RAES}$ from Eq.\ref{eq:sequence-autoencoder}. We found using these image- and sequence-autoencoders important for improving the latent space conditioning.
This combination of image-based and sequence-based losses assists in compressing the representation while ensuring intermediate representations remain informative.
The combined loss to train the  model is:
\begin{equation}
\label{eq:virl}
\begin{split}
&\mathcal{L}_{VIRL}(\textbf{\observationSeq}_{i}, \textbf{\observationSeq}_{p}, \changes{\textbf{\observationSeq}_{n}}, y; \encoderSymbol, \decoderSymbol, \omega, \rho) =   
 \underbrace{\lambda_{1} \mathcal{L}_{SNS}(\textbf{\observationSeq}_{i},\textbf{\observationSeq}_{p}, \changes{\textbf{\observationSeq}_{n}}, y; \encoderSymbol, \omega)}_{\text{Contrastive sequence loss (Eq3)}} + 
 \\
 & \underbrace{\lambda_{2} \Big[\frac{1}{T}\sum_{\ttime=0}^{T} \mathcal{L}_{SNI}(\textbf{\observationSeq}_{i,t}, \textbf{\observationSeq}_{p,t}, \changes{\textbf{\observationSeq}_{n,t}}, y; \encoderSymbol)\Big]}_{\text{Contrastive frame loss (Eq3)}} + \\
 & \underbrace{\lambda_{3} [\mathcal{L}_{RAES}( \textbf{\observationSeq}_{i}; \encoderSymbol, \decoderSymbol, \omega, \rho) +
\mathcal{L}_{RAES}(\textbf{\observationSeq}_{p}; \encoderSymbol, \decoderSymbol, \omega, \rho) \changes{+ \mathcal{L}_{RAES}(\textbf{\observationSeq}_{n}; \encoderSymbol, \decoderSymbol, \omega, \rho)]}}_{\text{Recurrent autoencoder loss (Eq.2)}} + \\
 & \underbrace{\lambda_{4} \Big[\frac{1}{T}\sum_{\ttime=0}^{T} \big[ \mathcal{L}_{VAEI}(\textbf{\observationSeq}_{i,t}; \encoderSymbol, \decoderSymbol) + \mathcal{L}_{VAEI}( \textbf{\observationSeq}_{p,t}; \encoderSymbol, \decoderSymbol) 
 \changes{ + \mathcal{L}_{VAEI}( \textbf{\observationSeq}_{n,t}; \encoderSymbol, \decoderSymbol)}\big] \Big].}_{\text{Variational autoencoder loss (individual frames) (Eq1)}}
\end{split}
\end{equation}
Where the relative weights of the different terms are $\lambda_{1:4} = \{ 0.7, 0.1, 0.1, 0.1\}$, the image encoder convnet is $\encoderSymbol$, the image decoder $\decoderSymbol$, the recurrent encoder $\omega$, and the recurrent decoder $\rho$. \changes{The weights for $\lambda$ are found by empirically evaluating \methodName over all environments from \autoref{subsection:sim-env}. Additional details on the hyperparameter search can be found in \autoref{sec:app-hyperparam-analysis}.}

\textbf{Reward Calculation}$\;$
\label{subsection:method-reward}
The model trained using the loss function described above is used to calculate the distance between two sequences of observations seen up to time $t$ as $\distance{\textbf{\observationSeq}^{e},\textbf{\observationSeq}^{a}; \phi, \omega} = ||f(\observation^{e}_{0:t}; \omega) - f(\observation^{a}_{0:t}; \omega)|| + ||f(\observation^{e}_{t}; \phi) - f(\observation^{a}_{t}, \phi)||$ and the reward as
$\reward(\observation^{e}_{0:t},\observation^{a}_{0:t}) = -\distance{\textbf{\observationSeq}^{e},\textbf{\observationSeq}^{a}; \phi, \omega}$.
During \ac{RL} training, we compute a distance given the sequence observed so far in the episode.
\newchanges{The sequence-based distance can model time-invariant distances, and the image-based distance can match the expert demonstration more precisely.}
In~\autoref{sec:experiments-ablation} we experimentally show the importance of each distance for imitation learning.

\begin{algorithm}[H]
	\caption{\textbf{\methodName}}
	\label{alg:VizImitation}
	\begin{algorithmic}[1]
		\State Initialize parameters \changes{$\modelParametersPolicy$, $\omega, \phi, \rho, \psi D \leftarrow \{\} $}
		\While{not done}
			\For{$i \in \{0, \ldots, N\}$} \Comment{Collect N trajectories}
				\State \notice{$\{\myState_{0}, \observation^{e}_{0}, \observation^{a}_{0}\} \leftarrow$} env.reset(), $\trajectory^{i} \leftarrow \{\}$ \Comment{\changes{Env returns state and observations, no reward}}
				\For{$\ttime \in \{0, \ldots, T\}$} \Comment{Collect a trajectory}
					\State $\action_{\ttime} \leftarrow \policySymbol(\cdot | \myState_{\ttime}, \modelParametersPolicy)$  \Comment{\changes{policy  produces an action given that agent state}}
					\State \notice{$\{\myState_{\ttime+1}, \observation^{e}_{\ttime+1}, \observation^{a}_{\ttime+1}\}  \leftarrow $} env.step($\action_{\ttime}$)		
					\State \changes{$\trajectory^{i}_{\ttime} \leftarrow \{\myState_{\ttime}, \observation^{e}_{\ttime}, \observation^{a}_{\ttime}, \action_{\ttime}\}$, $\textbf{\observationSeq}^{e}_{i,t} \leftarrow \observation^{e}_{\ttime}$, $\textbf{\observationSeq}^{a}_{i,t} \leftarrow \observation^{a}_{\ttime}$}  \Comment{\changes{store experience for later training}}
					\State{$\{\myState_{\ttime}, \observation^{e}_{\ttime}, \observation^{a}_{\ttime}\} \leftarrow \{\myState_{\ttime+1}, \observation^{e}_{\ttime+1}, \observation^{a}_{\ttime+1}\} $ } %
				\EndFor
				\State $D \leftarrow D \bigcup \{ \trajectory^{0}, \ldots, \trajectory^{N}\}$ \Comment{add trajectories to offline dataset}
				\State \changes{Perform $M$ batch updates $\distance{\cdot, \cdot}$ parameters $\omega, \phi, \rho, \psi$ using $D$ and \autoref{eq:virl}}
				\State \changes{$\textbf{\reward}^{0:N}_{0:T} \leftarrow -d(\textbf{\observationSeq}^{e}_{0:N,0:T}, \textbf{\observationSeq}^{a}_{0:N,0:T}| \omega, \phi)$} \Comment{\changes{compute rewards with updated $\phi, \omega$}}
				\State Update  $\modelParametersPolicy$ with $\{\{\trajectory^{0}, \textbf{\reward}^{0}\}, \ldots, \{\trajectory^{N}, \textbf{\reward}^{N}\}\}$  \Comment{\changes{RL policy update.}}
			\EndFor
		\EndWhile
	\end{algorithmic}
\end{algorithm}

\paragraph{Training the Model}
\label{subsection:method-training}       
Details of the algorithm used to train the \distanceMetricText and policy are in~\autoref{alg:VizImitation}.
We consider a variation on the typical \ac{RL} environment that produces three different outputs, two for the \agent and one for the \demonstrationText and no reward.
The first is the internal robot pose and link velocities, which we refer to as the state $\myState_{\ttime}$. 
The second and third are images of the \agent, or observation $\observation^{a}_{\ttime}$ and the \demonstrationText$\observation^{e}_{\ttime}$, shown in~\autoref{fig:overview-rewards}.
The images are used with the \distanceMetricText to compute the similarity between the \agent and the \demonstrationText.
We train the agent's policy using the \ac{TRPO} algorithm~\citep{TRPO}.
\changes{The policy uses the state $\myState_{\ttime}$ as input, which is easier to access than the 3rd person video generated by the agent during test time, and is trained online (line $15$) using the} learned distance function (line $14$). The use of off-policy training increases the \ac{LSTM}-based distance function's training efficiency. The off-policy training also allows us to train the distance function using data from other tasks to increase the robustness of the model while fine-tuning the current task, as we will describe in Section~\ref{sec:exp-ablation}.

\subsection{Unsupervised Data Labelling and Generation}
\label{subsection:data-augmentation}
To construct \textit{positive} and \textit{negative} pairs for training, we make use of time information 
and adversarial information. %
We use timing information where observations at similar times in the same sequence are often correlated, and observations at different times are less likely to be similar.
We use these ideas to provide labels for the positive and negative pairs to train the Siamese network.
Positive pairs are created by adding Gaussian noise with $\sigma = 0.05$ to the images in the sequence, duplicating, or shifting random frames of the sequences.
Negative pairs are created by shuffling, cropping or reversing one sequence.
\newchanges{\methodName will learn to decode these modified sequences which helps the model be robust to noise.}
Additionally, we include \textit{adversarial} pairs where positive pairs come from the same distribution, for example, two motions for the agent or two from the demonstration at different times. Negative pairs then include one from the expert and one from the agent. \newchanges{A combination of these augmentations are chosen randomly and applied to the current training batch and are not added to the replay buffer.} Additional details on how the shuffling, swapping, and the use of adversarial pairs are available in the \appendixx~\ref{sec:appendix-pos-neg-examples}.

\textbf{Data Augmentation}
\label{sec:data-augmentation}
We apply several data augmentation methods to produce additional data variation for training the \distanceMetricText.
Using methods analogous to the cropping and warping methods popular in computer vision~\citep{7005506} we randomly \textit{crop} sequences and randomly \textit{warp} the \demonstrationText timing.
The \textit{cropping} is performed by removing later portions of the training sequences during batch updates. These augmentations allow the distance function to learn a more general representation of motions, such as a walk even if the walking demonstration includes a single step or multiple steps at a different speed and pauses between steps.
This cropping is denoted as \ac{EESP} where the probability of cropping out a window in the sequence $\textbf{x}$ at $i$ is $p(i) = \frac{len(\textbf{x}) - i}{\sum i}$, increases the likelihood of cropping earlier in the sequence.
As the agent improves, the average length of each episode increases, and so too will the average length of the cropped window.
\changes{The motion \textit{warping} performs a type of scaling to the time dimension of the \demonstrationText. This is accomplished by creating a continuous function of the \demonstrationText so we can sample the motion at different speeds $\hat{\observationSeq} \leftarrow f(\observationSeq, \delta)$. For example, if the motion is replayed at $\delta=0.5$ the motion $\hat{\observationSeq}$ will be twice as long as $\observationSeq$. Linear interpolation is used to fill in information between frames.}
This allows for training the distance function to recognize that, for example, a walking motion with one step in it and another with two steps are still examples of a walk. 
Last, we use \ac{RSI}~\citep{2018-TOG-deepMimic}, where we generate the initial state of the \agent and \expert randomly from the \demonstrationText.
With this property, the environment functions as a form of memory replay.
The environment allows the agent to go back to random points in the \demonstrationText as if replaying a remembered demonstration and collect new data from that point in the demonstration. The experiments in~\refSection{sec:experiments-ablation} show the importance of these augmentation methods in terms of improving the robustness of the learned comparator and resulting policy.
\section{Experiments, Results and Analysis}
\label{sec:results}
\label{subsection:sim-env}
\begin{wrapfigure}{r}{0.5\linewidth} %
\vspace{-0.2cm}
\centering 
\adjustbox{trim={.425\width} {.2\height} {0.445\width} {.25\height},clip}{\includegraphics[width=0.75\linewidth]{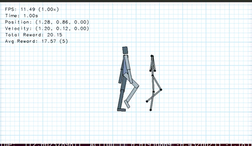}}
\hspace{.25cm}
\adjustbox{trim={0.0\width} {0.0\height} {0.0\width} {0.0\height},clip}{\includegraphics[width=0.3\linewidth]{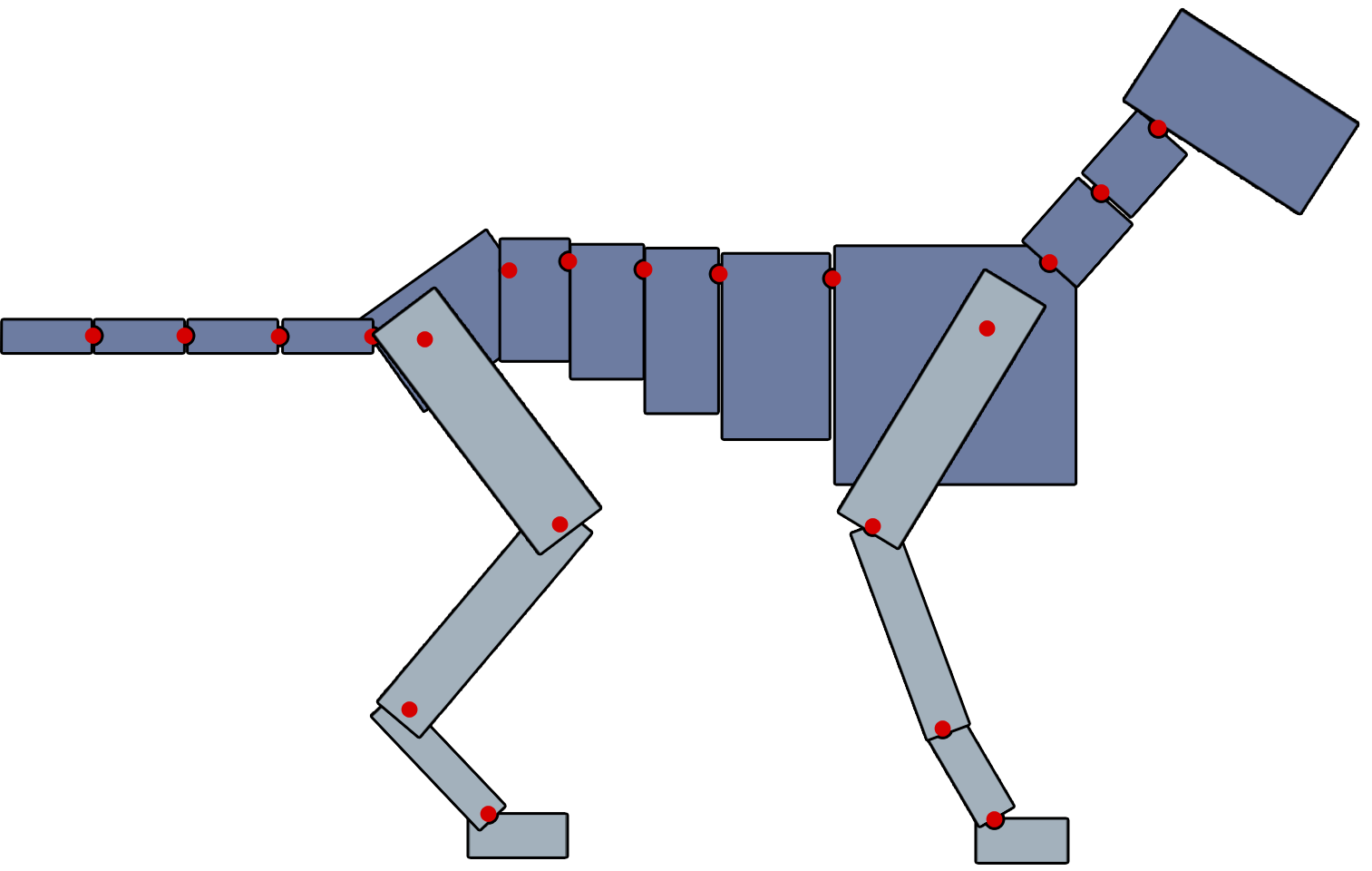} } 
\adjustbox{trim={0.0\width} {0.0\height} {0.0\width} {0.0\height},clip}{\includegraphics[width=0.35\linewidth]{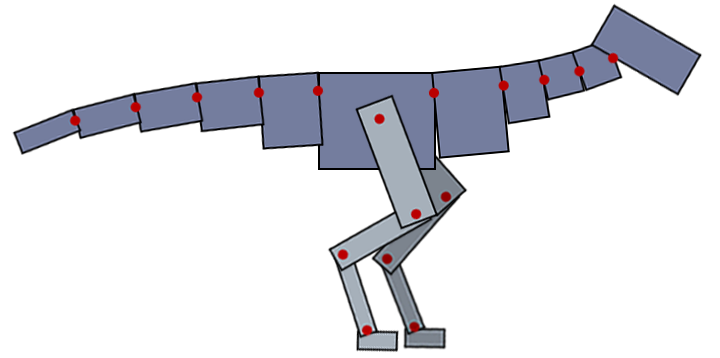} }
\caption{ 
Frames from the \humanoidTwoD, dog2d, and raptor2d environments used in the experiments.
}
\label{fig:mixed-environemnts}
\vspace{-0.2cm}
\end{wrapfigure}

\begin{figure*}[b!]%
\centering

\adjustbox{trim={.425\width} {.22\height} {0.445\width} {.29\height},clip}{\includegraphics[width=0.5\linewidth]{images/humanoid2d/DDPG/walking/humanoid2d_ddpg_viz_imitation-3.png}}
\hspace{.25cm}
\adjustbox{trim={.425\width} {.22\height} {0.445\width} {.29\height},clip}{\includegraphics[width=0.5\linewidth]{images/humanoid2d/DDPG/walking/humanoid2d_ddpg_viz_imitation-3.png}}
\adjustbox{trim={.425\width} {.22\height} {0.445\width} {.29\height},clip}{\includegraphics[width=0.5\linewidth]{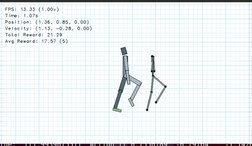} }
\adjustbox{trim={.425\width} {.22\height} {0.445\width} {.29\height},clip}{\includegraphics[width=0.5\linewidth]{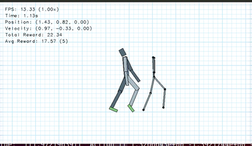}}
\adjustbox{trim={.425\width} {.22\height} {0.445\width} {.29\height},clip}{\includegraphics[width=0.5\linewidth]{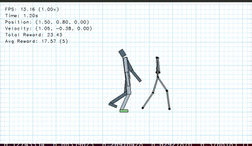} }
\adjustbox{trim={.425\width} {.22\height} {0.445\width} {.29\height},clip}{\includegraphics[width=0.5\linewidth]{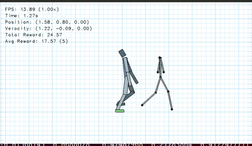} }
\adjustbox{trim={.425\width} {.22\height} {0.445\width} {.29\height},clip}{\includegraphics[width=0.5\linewidth]{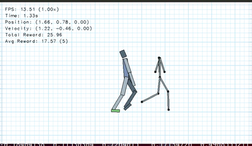} }
\adjustbox{trim={.425\width} {.22\height} {0.445\width} {.29\height},clip}{\includegraphics[width=0.5\linewidth]{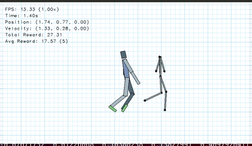} }
\adjustbox{trim={.425\width} {.22\height} {0.445\width} {.29\height},clip}{\includegraphics[width=0.5\linewidth]{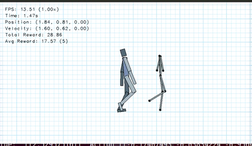} }
\adjustbox{trim={.425\width} {.22\height} {0.445\width} {.29\height},clip}{\includegraphics[width=0.5\linewidth]{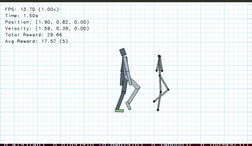} }
\adjustbox{trim={.425\width} {.22\height} {0.445\width} {.29\height},clip}{\includegraphics[width=0.5\linewidth]{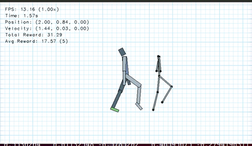} }
\adjustbox{trim={.425\width} {.22\height} {0.445\width} {.29\height},clip}{\includegraphics[width=0.5\linewidth]{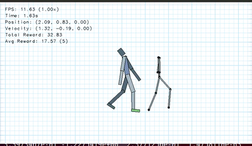} } \\

\adjustbox{trim={.4\width} {.20\height} {0.4\width} {.25\height},clip}{\includegraphics[width=0.39\linewidth]{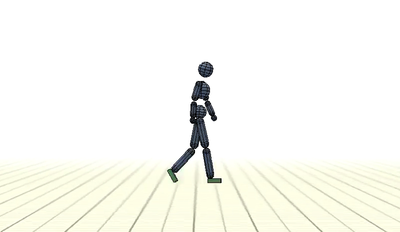}}
\adjustbox{trim={.4\width} {.20\height} {0.4\width} {.25\height},clip}{\includegraphics[width=0.39\linewidth]{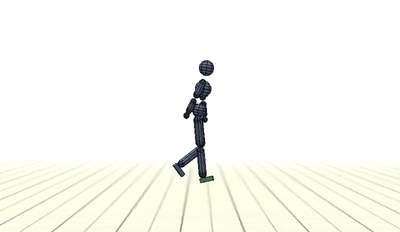}}
\adjustbox{trim={.4\width} {.20\height} {0.4\width} {.25\height},clip}{\includegraphics[width=0.39\linewidth]{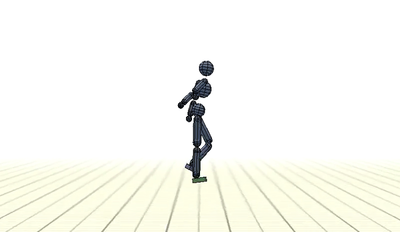}}
\adjustbox{trim={.4\width} {.20\height} {0.4\width} {.25\height},clip}{\includegraphics[width=0.39\linewidth]{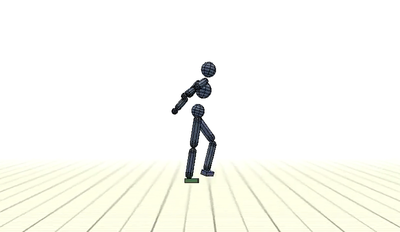}}
\adjustbox{trim={.4\width} {.20\height} {0.4\width} {.25\height},clip}{\includegraphics[width=0.39\linewidth]{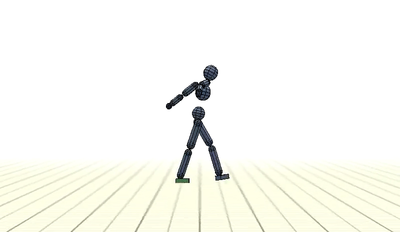}}
\adjustbox{trim={.4\width} {.20\height} {0.4\width} {.25\height},clip}{\includegraphics[width=0.39\linewidth]{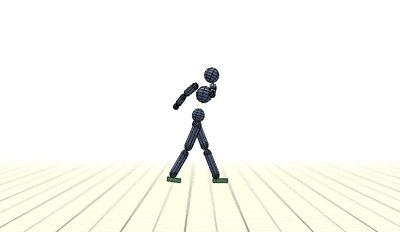}}
\adjustbox{trim={.4\width} {.20\height} {0.4\width} {.25\height},clip}{\includegraphics[width=0.39\linewidth]{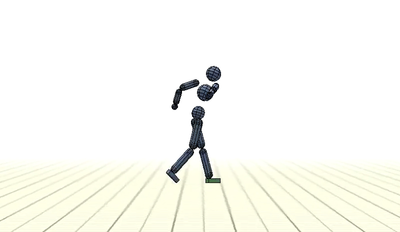}}
\adjustbox{trim={.4\width} {.20\height} {0.4\width} {.25\height},clip}{\includegraphics[width=0.39\linewidth]{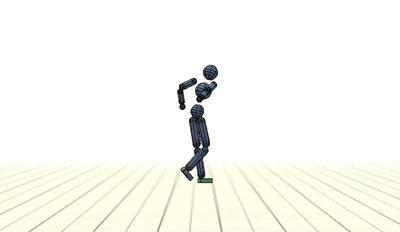}}
\adjustbox{trim={.4\width} {.20\height} {0.4\width} {.25\height},clip}{\includegraphics[width=0.39\linewidth]{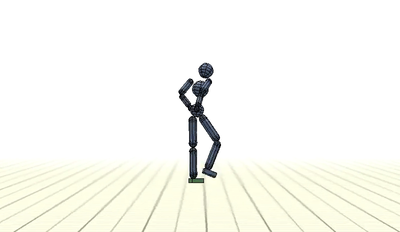}}
\adjustbox{trim={.4\width} {.20\height} {0.4\width} {.25\height},clip}{\includegraphics[width=0.39\linewidth]{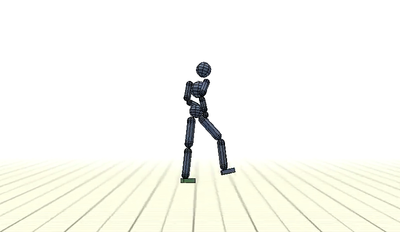}}
\adjustbox{trim={.4\width} {.20\height} {0.4\width} {.25\height},clip}{\includegraphics[width=0.39\linewidth]{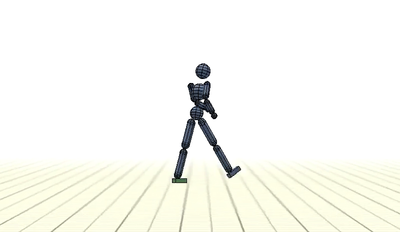}}\\

\caption{ 
Frames of the \agent's motion after training on \humanoidTwoD and \humanoidThreeD walking. Additionally, the dog, raptor, zombie walk, run and jumping policy can be found on the project website: ~\href{\videoLink}{\videoLink}.
}
\label{fig:humanoid3d-results}

\end{figure*}

We evaluate \methodName compared to previous methods in terms of sample efficiency and task-solving capability. The comparison is over a collection of different simulation environments.
In these simulated robotics environments, we task the \agent with imitating a given reference video demonstration.
Each simulation environment provides a hard-coded reward function based on the \agent's pose to evaluate the policy quality independently.
The imitation environments include challenging and dynamic tasks for humanoid, dog and raptor robots.
Some example tasks are running, jumping, trotting, and walking, shown in~\refFigure{fig:mixed-environemnts} and~\refFigure{fig:humanoid3d-results}.
The \demonstrationText $\textbf{\observation}^{e}_{\ttime}$ the \agent uses for visual imitation learning is produced from a clip of motion capture data for each task.
The motion capture data animates a kinematically controlled robot in the simulation for capturing video. Because the demonstration is kinematically generated, the agent also needs to learn how to bridge the gap between the different dynamics in the demonstration that may be impossible to reproduce exactly. 
We convert the images captured from the simulation to $48 \times 48$ grey-scale images. Third-person image data is often not available during test time; therefore, the \agent's policy instead receives the environment state as the link distances and velocities relative to the robot's \ac{COM}.

\textbf{2D Imitation Tasks}$\;$
The first group of evaluation environments contain a set of agents with different morphologies.
In Figure~\ref{fig:mixed-environemnts} we show images from the 2D humanoid, dog and raptor environment. In these environments the rendering and simulation is in 2D, reducing the complexity of the control system and dynamics, and allowing for faster training times.

\textbf{3D Imitation Tasks}$\;$
For further evaluation, we compare the performance on 3D humanoid and two quadrupedal robot simulators used for Sim2Real research, the Laikago~\citep{RoboImitationPeng20} and Pupper~\citep{pupperrobot}. 
The \humanoidThreeD environment has multiple tasks that can be solved and used to generate data from other tasks for training the \distanceMetricText. These tasks include: walking, running, jogging, front-flips, back-flips, dancing, jumping, punching and kicking). 
To perform the data augmentation described in Section~\ref{sec:data-augmentation} we also construct data from a modified version of each task with a randomly generated playback speed modifier \changes{$\delta \in [0.5, 2.0] $} e.g. walking-dynamic-speed, which warps the \demonstrationText timing. 
This additional data provides a richer understanding of distances in space and time with the \distanceMetricText.
As we will show later in this section, \methodName learns policies that produce similar behaviour to the \demonstrationText across these diverse tasks.
We show example trajectories from the learned policies in~\refFigure{fig:humanoid3d-results} and in the supplemental~\href{\videoLink}{Video}.
It takes $5-7$ days to train each policy in these results on a $16$ core machine with an Nvidia GTX1080 GPU.

\paragraph{Comparison Methods} We compare \methodName to two baselines that learn distances in observation space. The first is 
\ac{GAIfO}~\citep{torabi2018generative} that trains a \ac{GAN} to differentiate between images from the \demonstrationText and images from the \agent.
The other is \ac{TCN}, an image-to-image only siamese model~\citep{DBLP:journals/corr/abs-1807-04742}. These methods have been used to perform types of imitation from observation before. However, as we will see, they either require a significant amount of data to train or result in lower-quality reward functions and, as a result, lower-quality policies. 

\subsection{Learning Performance}
\label{sec:experiments-learning-performance}
\begin{figure*}[t!]
\centering
\subcaptionbox{\label{fig:humanoid2d-compare} humanoid2d walk}{ \includegraphics[trim={0.0cm 0.0cm 0.0cm 0.0cm},clip,width=0.31\linewidth]{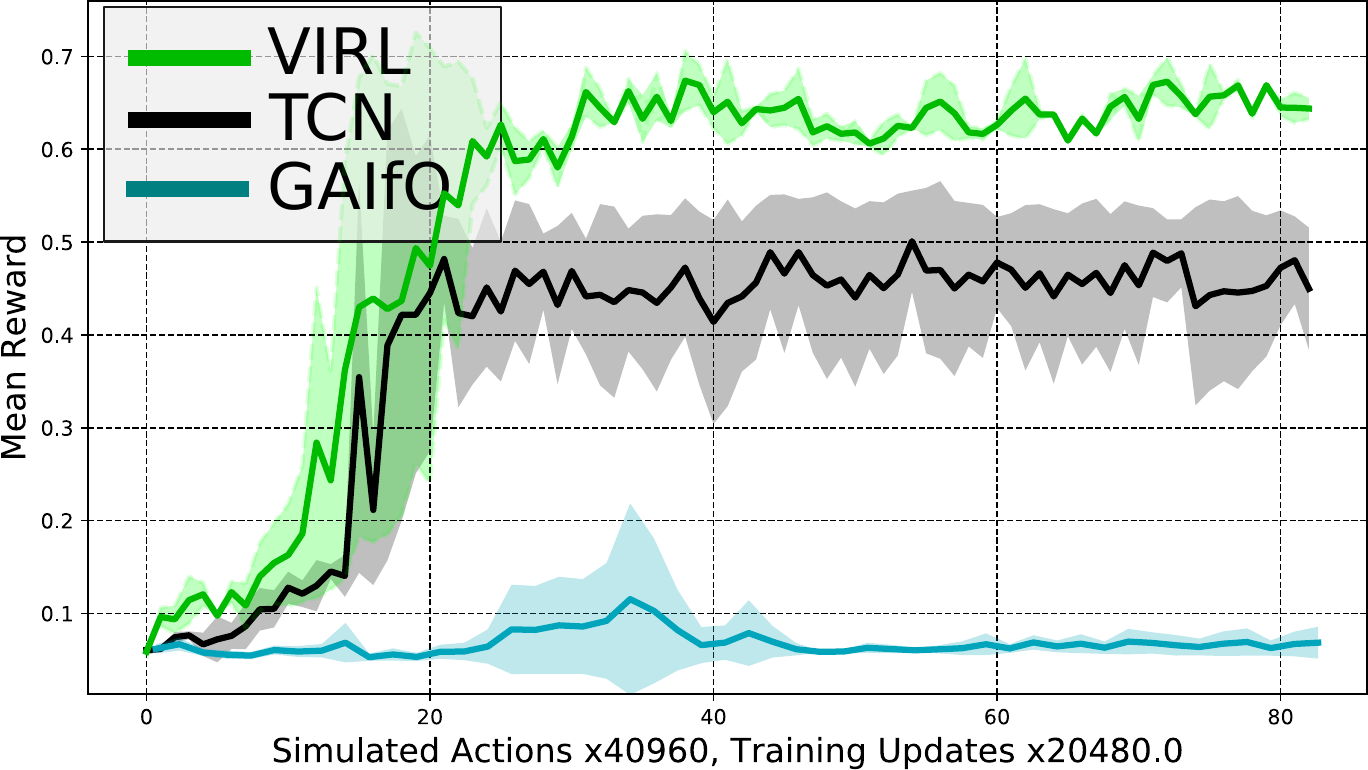}}
\subcaptionbox{\label{fig:dog2d-compare}  dog2d}{ \includegraphics[trim={0.0cm 0.0cm 0.0cm 0.0cm},clip,width=0.31\linewidth]{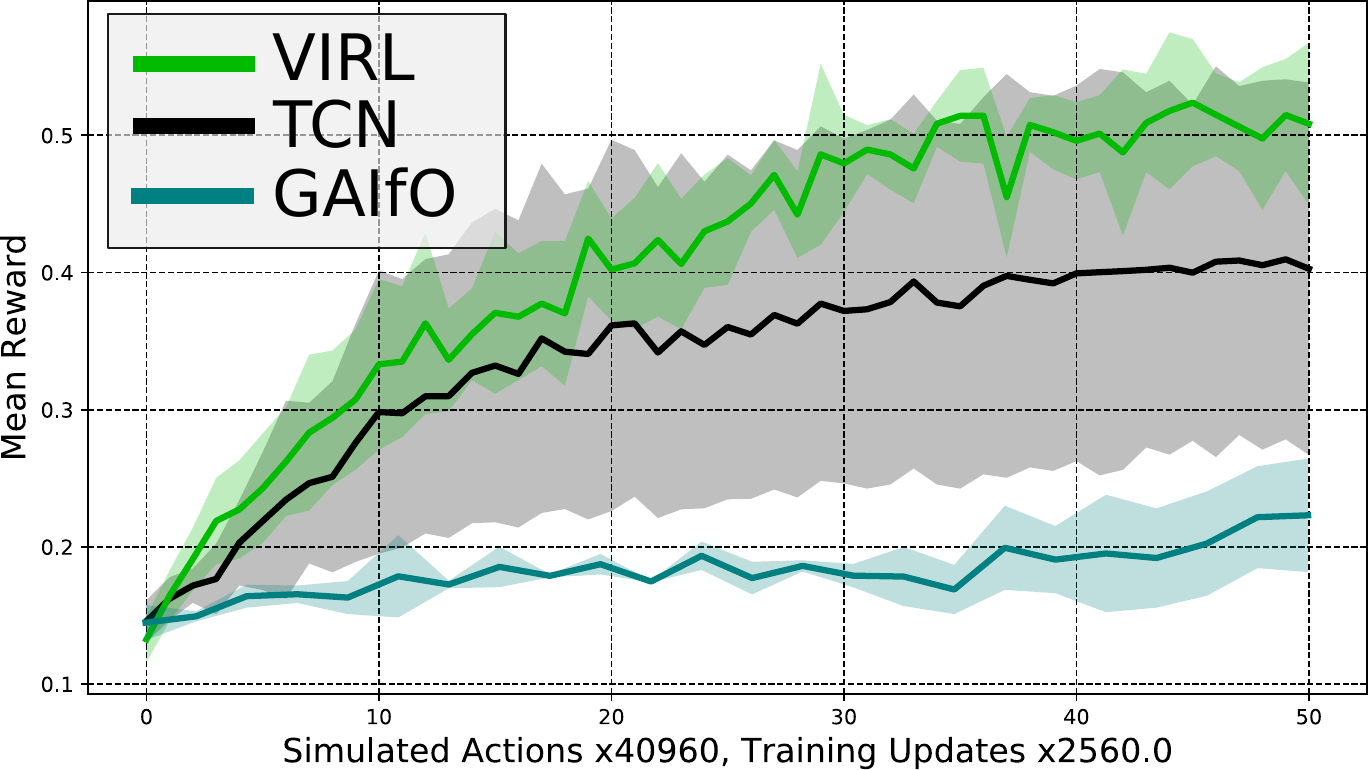}} 
\subcaptionbox{\label{fig:raptor2d}  raptor2d}{ \includegraphics[trim={0.0cm 0.0cm 0.0cm 0.0cm},clip,width=0.31\linewidth]{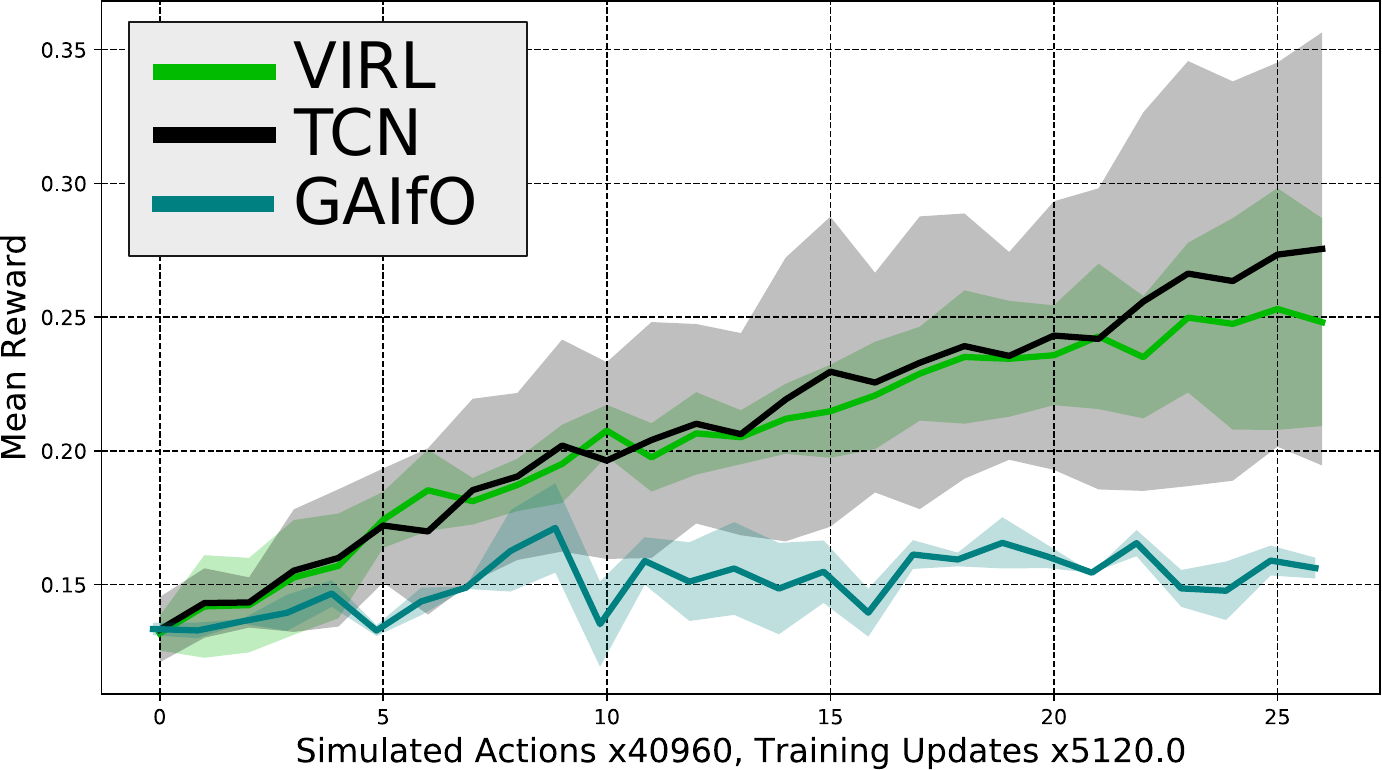}}
\caption{
Comparisons of \methodName, \ac{TCN} and \ac{GAIfO} with (a) the humanoid2d, (b) the 2D dog agent, and (c) the 2D raptor agent. \ac{GAIfO} struggles to show improvement on these tasks. \ac{TCN} does make progress on these imitation tasks but the performance is not as good as \methodName.
The results show average performance over $4$ randomly seeded policy training simulations.
}
\label{fig:humanoid2d-rl-compare-old-cd}
\vspace{-0.25cm}
\end{figure*}

In Figure \ref{fig:humanoid2d-rl-compare-old-cd} we present results across different agent types including the 2D walking humanoid in Section~\ref{fig:humanoid2d-compare}, a 2D trotting dog in ~\refFigure{fig:dog2d-compare}, and a 2D Raptor in ~\refFigure{fig:raptor2d}. We compare \methodName directly with both \ac{GAIfO} and \ac{TCN} -- the strongest comparable prior method.
\methodName is able to provide a denser reward signal due to the LSTM-based distance that can express similarity between motion styles.
While \ac{TCN} is simpler, not using an LSTM model, it does not provide as rich of a reward signal for the agent to learn from. As a result, the learned reward can often be more sparse, slowing down learning.
Similarly, \ac{GAIfO} has difficulty learning a robust and smooth distance function with the little \demonstrationText data available. This leads to very jerky motion or \agent's that stand still, matching the average pose of the demonstration, both of which are contained in the demonstration distribution but do not capture sequential-temporal-behaviour well.  

\begin{figure*}[b!]
\centering
\subcaptionbox{\label{fig:vsTCN-walk} \small Walking  }{ \includegraphics[trim={0.0cm 0.0cm 0.0cm 0.0cm},clip,width=0.45\linewidth]{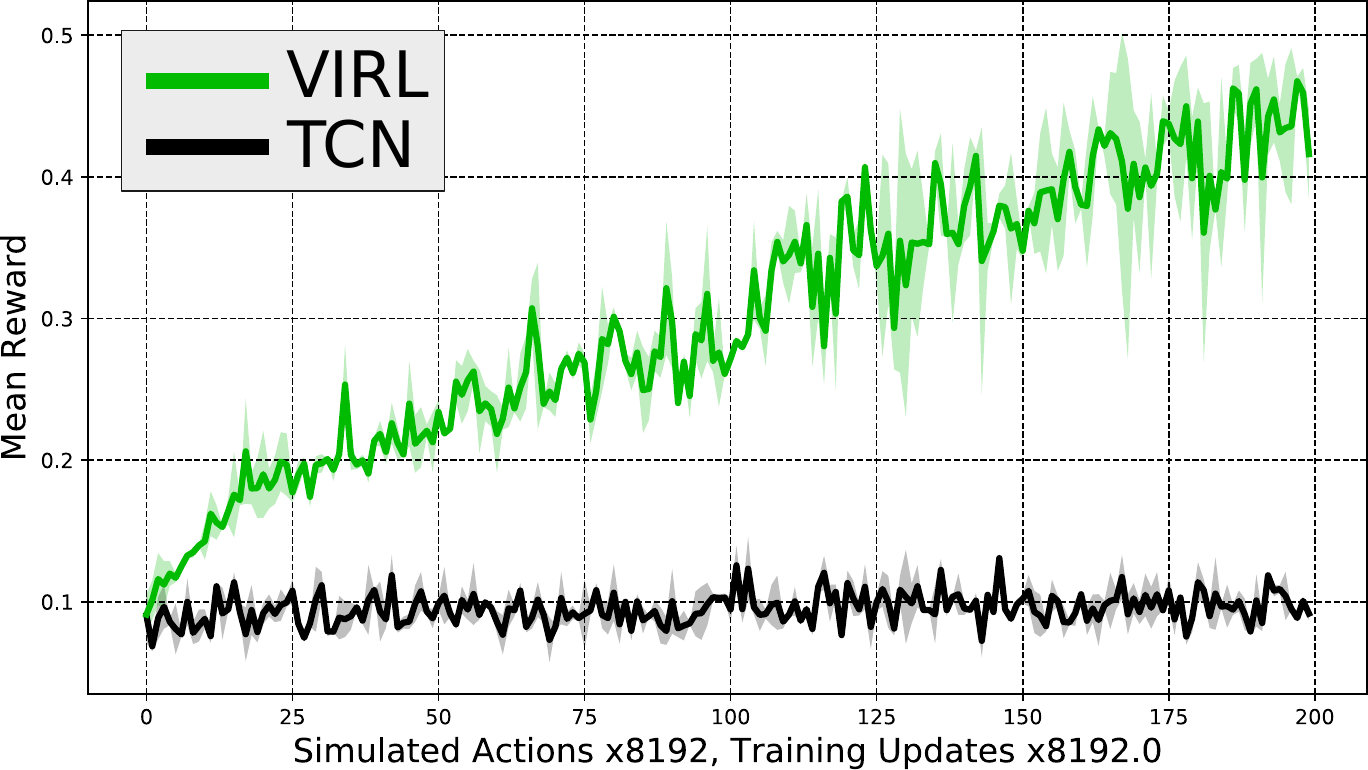}}
\subcaptionbox{\label{fig:vsTCN-zombiewalk} ZombieWalk}{ \includegraphics[trim={0.0cm 0.0cm 0.0cm 0.0cm},clip,width=0.45\linewidth]{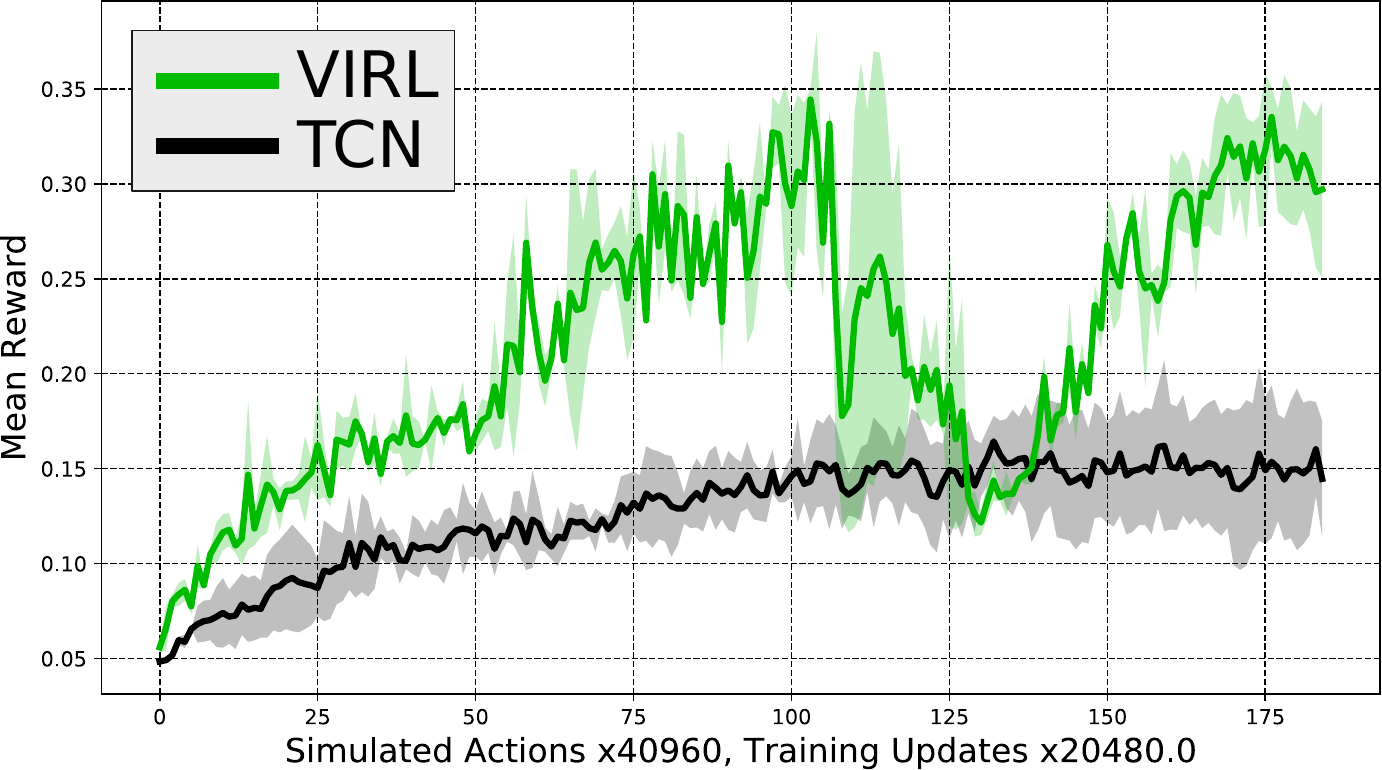}}
\subcaptionbox{\label{fig:vsTCN-run} \small Running}{ \includegraphics[trim={0.0cm 0.0cm 0.0cm 0.0cm},clip,width=0.45\linewidth]{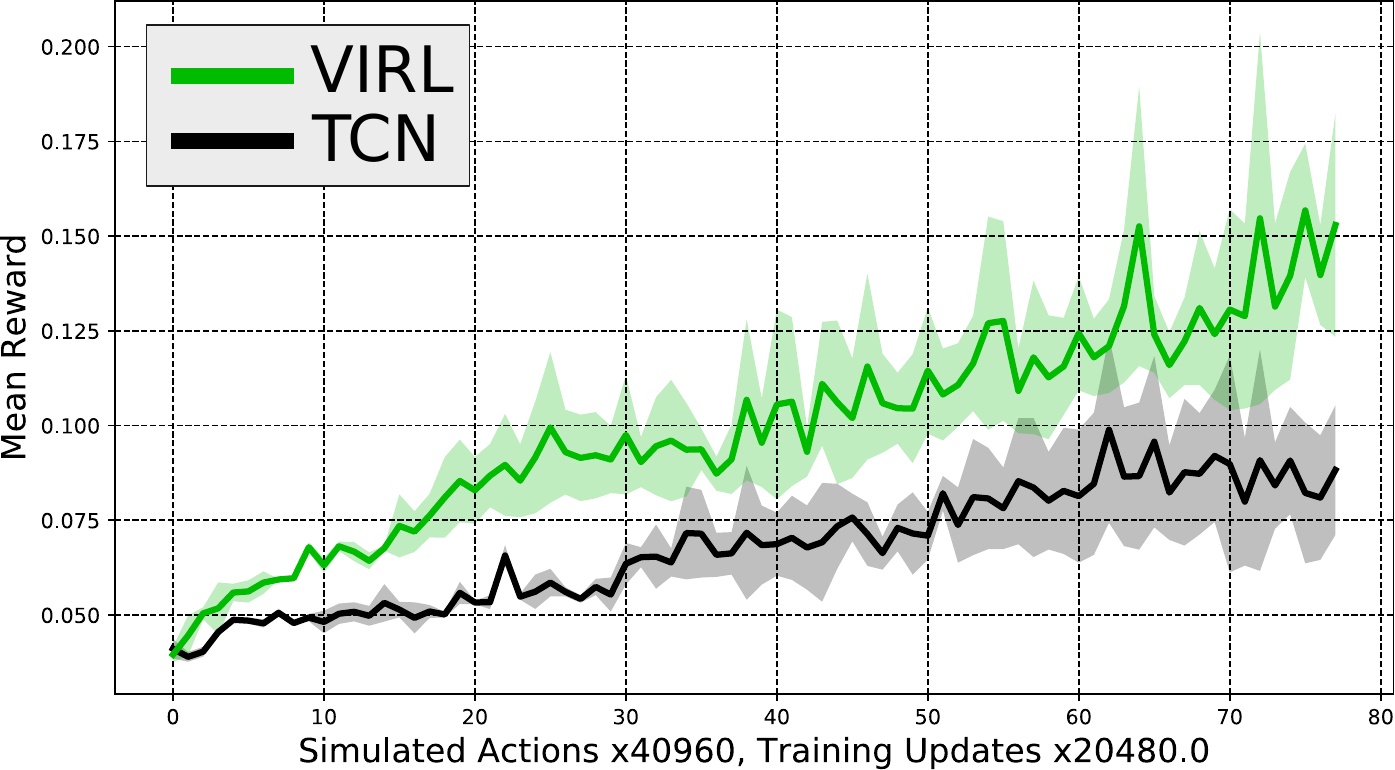}}
\subcaptionbox{\label{fig:vsTCN-jump}Jumping}{ \includegraphics[trim={0.0cm 0.0cm 0.0cm 0.0cm},clip,width=0.45\linewidth]{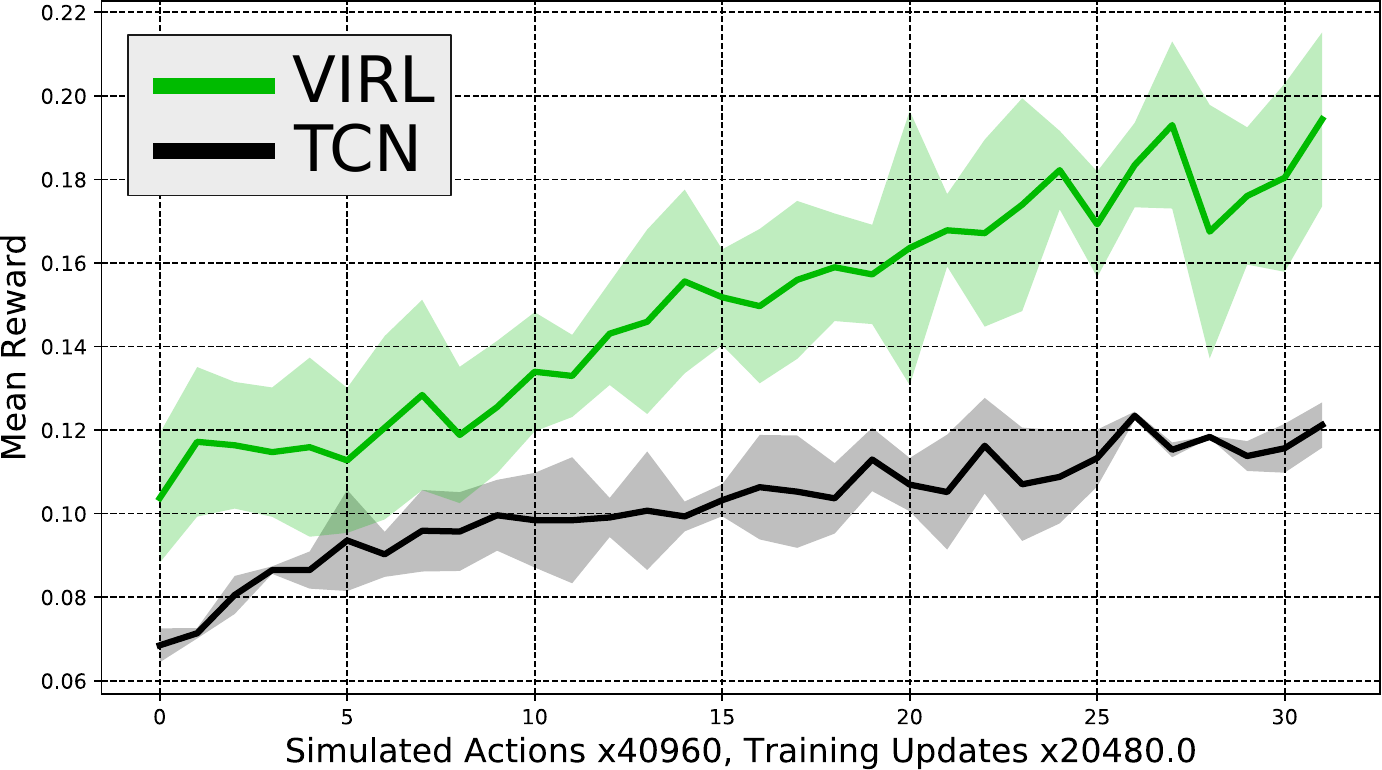}}
\caption{
3D Humanoid motion imitation experiments: Comparisons of \methodName with \ac{TCN}s. These results show the average performance over $4$ randomly seeded policy training simulations.
}
\label{fig:humanoid2d-rl-compare-old-eh}
\vspace{-0.25cm}
\end{figure*}

In Figure \ref{fig:humanoid2d-rl-compare-old-eh} we compare \methodName with \ac{TCN}, across many different and more challenging \humanoidThreeD tasks in~\refFigure{fig:humanoid2d-rl-compare-old-eh}(a-d). 
Across these experiments, we observe that \methodName learns faster and produces higher value policies.
In particular, we find that \methodName does very well compared to TCNs, which represents the strongest prior approach capable of performing this task of which we are aware. The \humanoidThreeD tasks are particularly challenging as they have high control dimensionality causing the agent to deviate from the desired imitation behaviour easily. These tasks also contain higher levels of partial observability compared to the 2D experiments. In these environments, the temporal distance in \methodName provides a crucial additional reward signal that helps the agent match the style of the motion early on in training despite the partial information observations.

\subsection{Analysis and Ablation}
\label{sec:experiments-ablation}
\begin{figure*}[h!]
\centering
\subcaptionbox{\label{fig:humanoid2d-embedding-humanoid3d}\ac{t-SNE} embedding (\humanoidThreeD)}{ \includegraphics[trim={0.0cm 0.0cm 0.0cm 0.0cm},clip,width=0.35\linewidth]{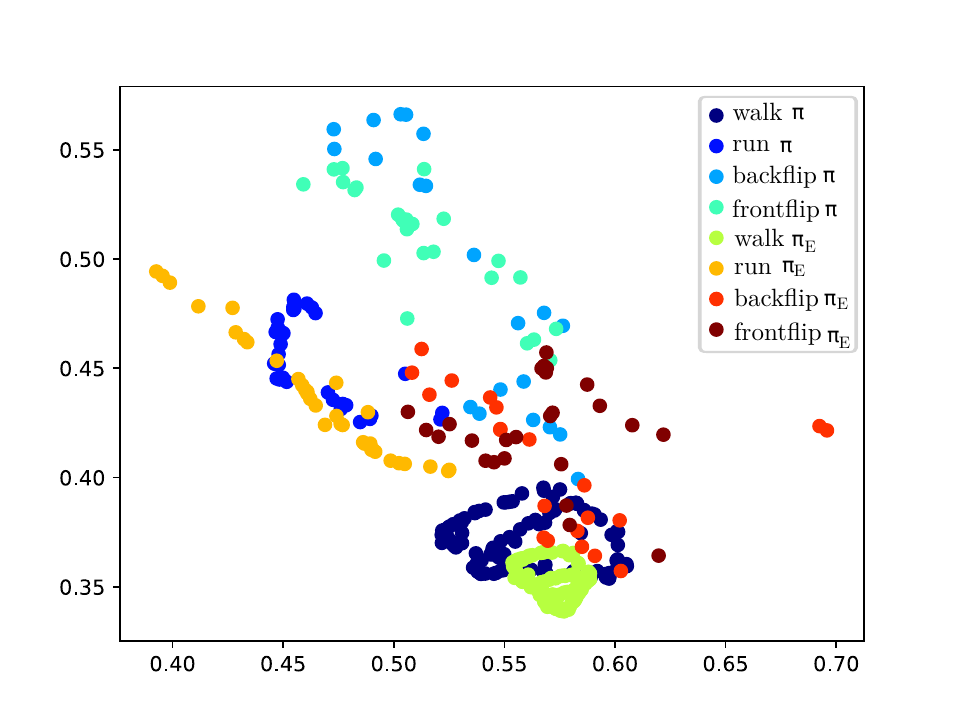}}
\subcaptionbox{\label{fig:spatial-vs-temporal} Distance function ablation}{ \includegraphics[trim={0.0cm 0.0cm 0.0cm 0.0cm},clip,width=0.45\linewidth]{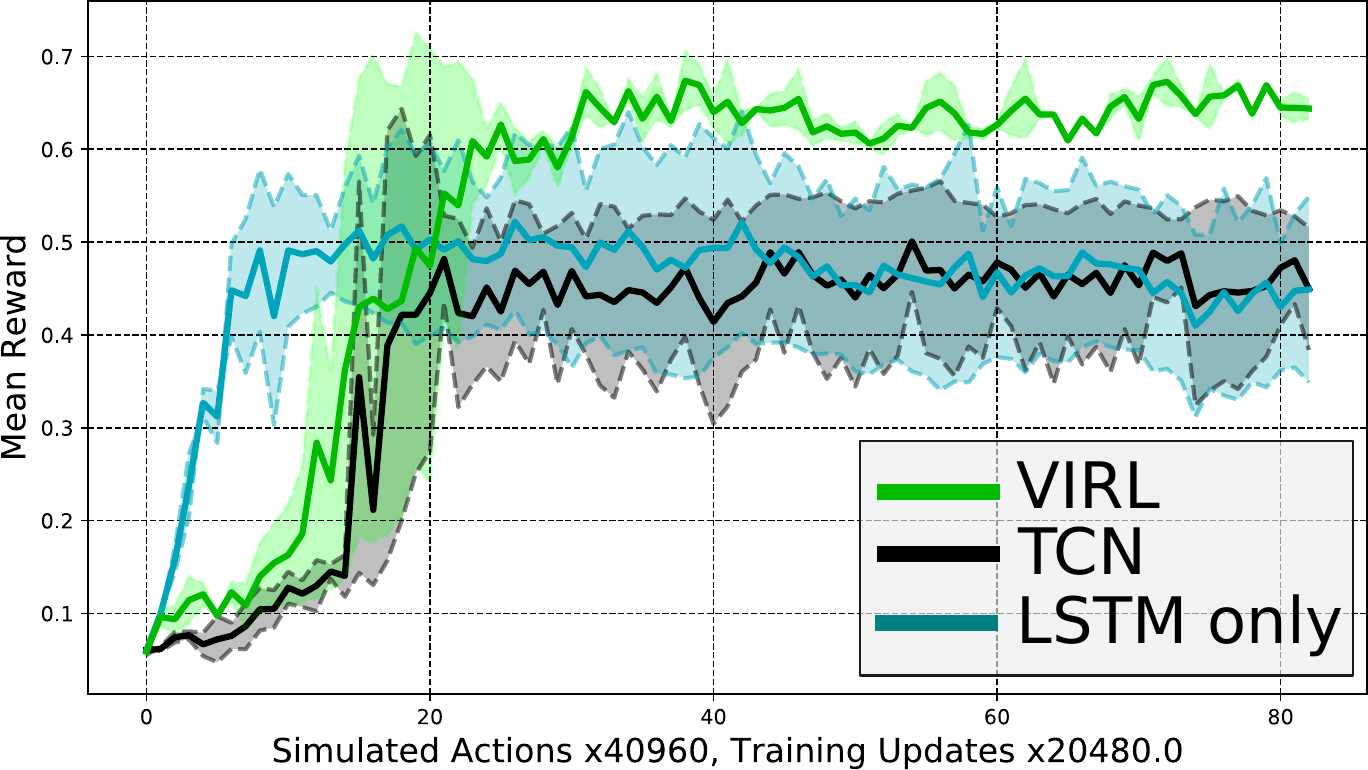}}
\caption{(a) t-SNE embedding that shows \methodName's learned distance landscape. (b) Comparison between combining different learned distance models. The combination of spatial and temporal distances leads to the best result (\methodName).
}
\label{fig:humanoid2d-rl-compare-old-ab}
\vspace{-0.25cm}
\end{figure*}

\textbf{Sequence Encoding}
Using the learned sequence encoder, we compute the encodings across a collection of different motions and create a \ac{t-SNE} embedding of the encodings~\citep{maaten2008visualizing}.
In~\refFigure{fig:humanoid2d-embedding-humanoid3d} we plot motions both generated from the learned policy $\policySymbol$ and the expert trajectories $\policySymbol_{E}$.
Overlaps in specific areas of the space for similar classes across learned $\policySymbol$ and expert $\policySymbol_{E}$ data indicate a well-formed distance metric that does not separate \expert and \agent examples. %
There is also a separation between motion classes in the data, and the cyclic nature of the walking cycle is visible.

\textbf{Ablation}$\;$
\label{sec:exp-ablation}
In~\refFigure{fig:spatial-vs-temporal} we compare the importance of the spatial distance \changes{ $||f(\observation^{e}_{t}; \phi) - f(\observation^{a}_{t}, \phi)||^2$ using the image encoder $\phi$ and temporal-LSTM-distance $||f(\observation^{e}_{0:t}; \omega) - f(\observation^{a}_{0:t}; \omega)||^2$ using the sequence encoder $\omega$ of \methodName}.
Using the recurrent representation alone (LSTM only) allows learning to progress quickly but can lead to difficulties in informing the policy on how to match the desired demonstration more precisely. On the other hand, using only the encoding between single frames as is done with TCN, slows learning due to little reward when the agent quickly becomes out-of-sync with the demonstration behaviour. 
The best result is achieved by combining the representations from these two models (\methodName).

\begin{figure*}[tb]
\centering
\subcaptionbox{\label{fig:ablation-humanoid3d} \small 3D Walking Ablation Analysis  }{ \includegraphics[trim={0.0cm 0.0cm 0.0cm 0.0cm},clip,width=0.45\linewidth]{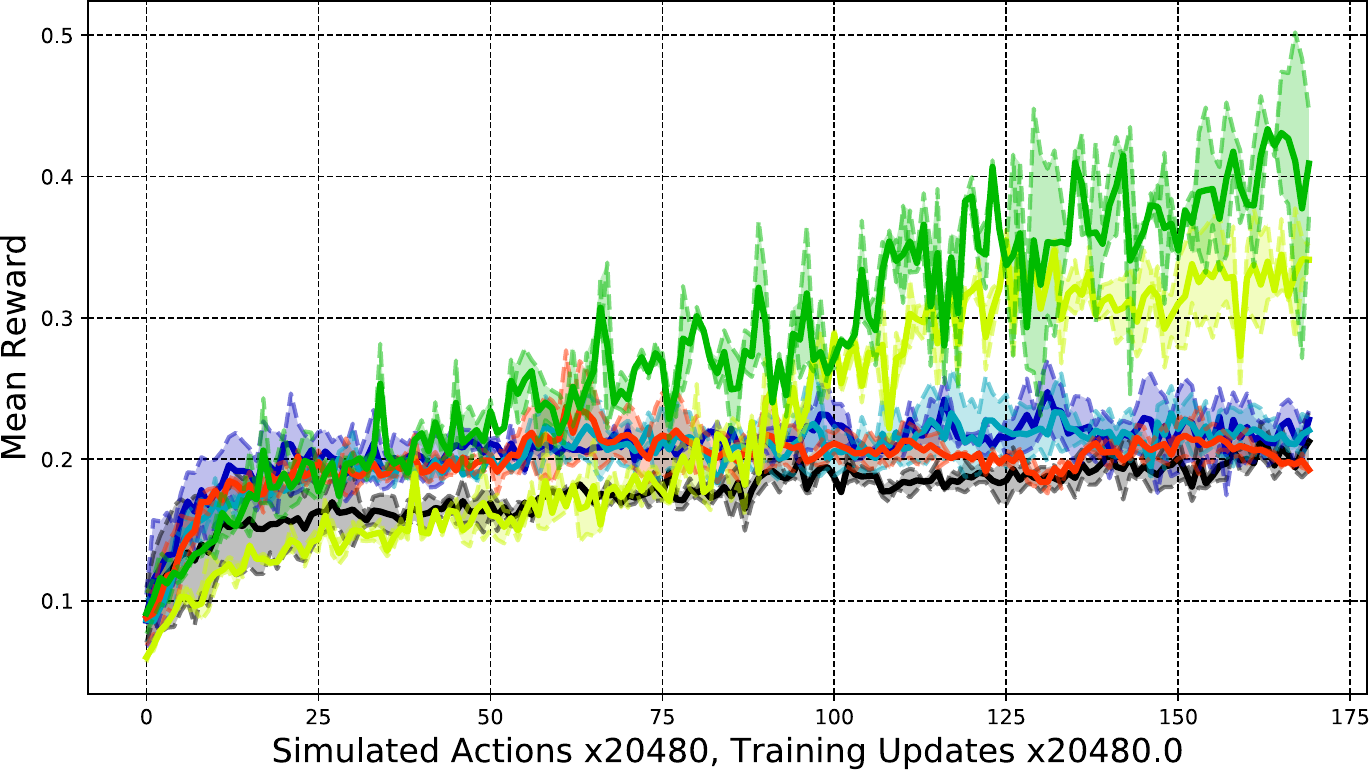}}
\subcaptionbox{\label{fig:ablation-fd-humanoid3d} \small Distance Metrics  }{ \includegraphics[trim={0.0cm 0.0cm 0.0cm 0.0cm},clip,width=0.45\linewidth]{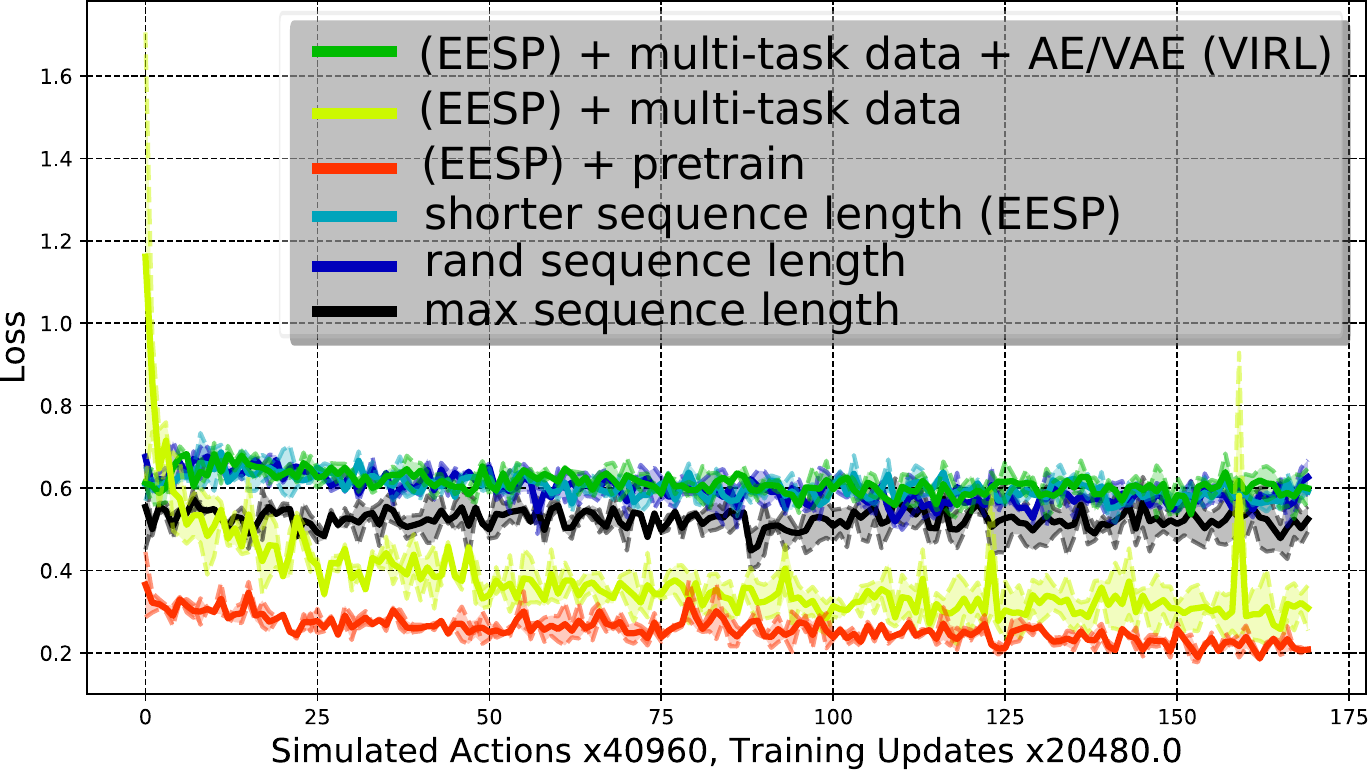}}
\caption{ 
(a) Ablation analysis of \methodName on the humanoid3D environment showing the mean reward over $5$ random seeds. The legend is the same as (b), where we examine the impact on our loss under the different distance metrics resulting from the ablation analysis. We find that including multi-task data (only available for the humanoid3D) and both the \ac{VAE} and recurrent \ac{AE} losses provide the most high-performing models. 
}
\label{fig:humanoid3d-hyperparam-analysis-old-ab}
\vspace{-0.25cm}
\end{figure*}

\textbf{Data augmentation comparisons.}
We conduct ablation studies for learning policies for 3D humanoid control in~\refFigure{fig:ablation-humanoid3d} and~\ref{fig:ablation-fd-humanoid3d}. We compare the effects of data augmentation methods, network models and the use of additional data from other tasks to train the siamese network ($24$ additional tasks such as back-flips, see appendix \ref{sec:appendix-data} for more details on these tasks). We also compared using different length sequences for training, shorter (where the probability of the length decays linearly), uniform random and max length available. For these more complex and challenging three-dimensional humanoids (humanoid3d) control problems, the data augmentation methods, including \ac{EESP}, increase average policy quality marginally compared to the importance of using multi-task data, this is likely related to the increased partial observably of these tasks. However, the addition of the auto-encoding losses to \methodName results in the quickest learning and highest value policies.

\begin{figure*}[tb]
\centering
\subcaptionbox{\label{fig:lstm-AE} ZombieWalk, LSTM }{ \includegraphics[trim={0.0cm 0.0cm 0.0cm 0.0cm},clip,width=0.45\linewidth]{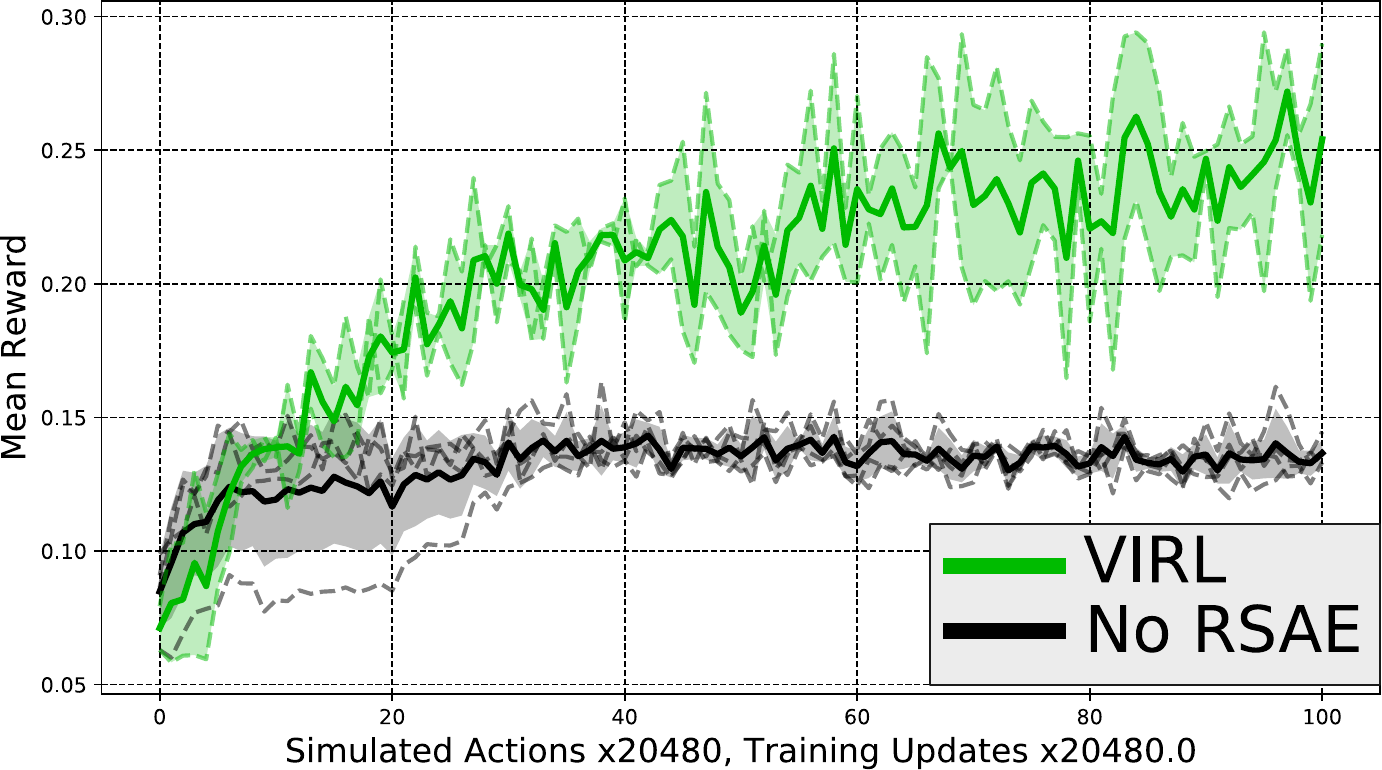}}
\subcaptionbox{\label{fig:run-multitask-data}Running, MultiTask}{ \includegraphics[trim={0.0cm 0.0cm 0.0cm 0.0cm},clip,width=0.45\linewidth]{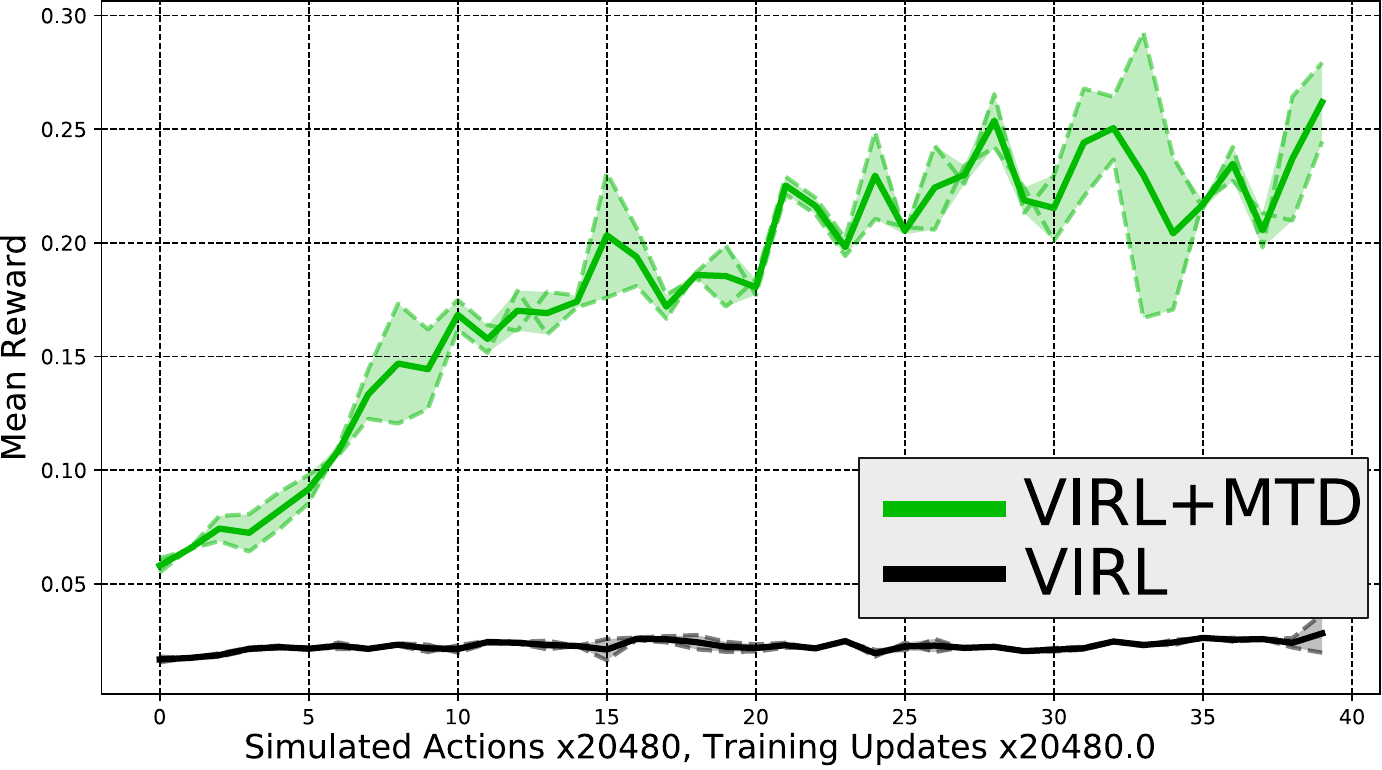}}
\caption{ 
(a) Ablating the recurrent autoencoder from \methodName (i.e. No RSAE in the figure) impairs the ability to learn how to walk like a Zombie. (b) 
Here, further we see how multi-task training data helps learn better policies for running (away from Zombies if desired). 
}
\label{fig:humanoid3d-hyperparam-analysis-old-cd}
\vspace{-0.25cm}
\end{figure*}

The experiment in~\refFigure{fig:lstm-AE} highlights the improvement to \methodName afforded by the \ac{RSAE} model component from Eq.~\ref{eq:virl} that forces the encoding to contain enough information to decode the video sequence. While the experiment in~\refFigure{fig:run-multitask-data} shows the dramatic improvement achieved when we include offline multi-task data for training the distance function. These methods are combined together to provide substantial learning performance increases across environments for \methodName.
Further analysis is available in the \appendixx, including additional comparison with \ac{TCN} in~\refFigure{fig:humanoid2d-hyperparam-analysis}(a-b) and details on training the distance model.

\begin{SCfigure}
\centering
\includegraphics[trim={18.0cm 15.0cm 18.0cm 14.0cm},clip,width=0.25\linewidth]{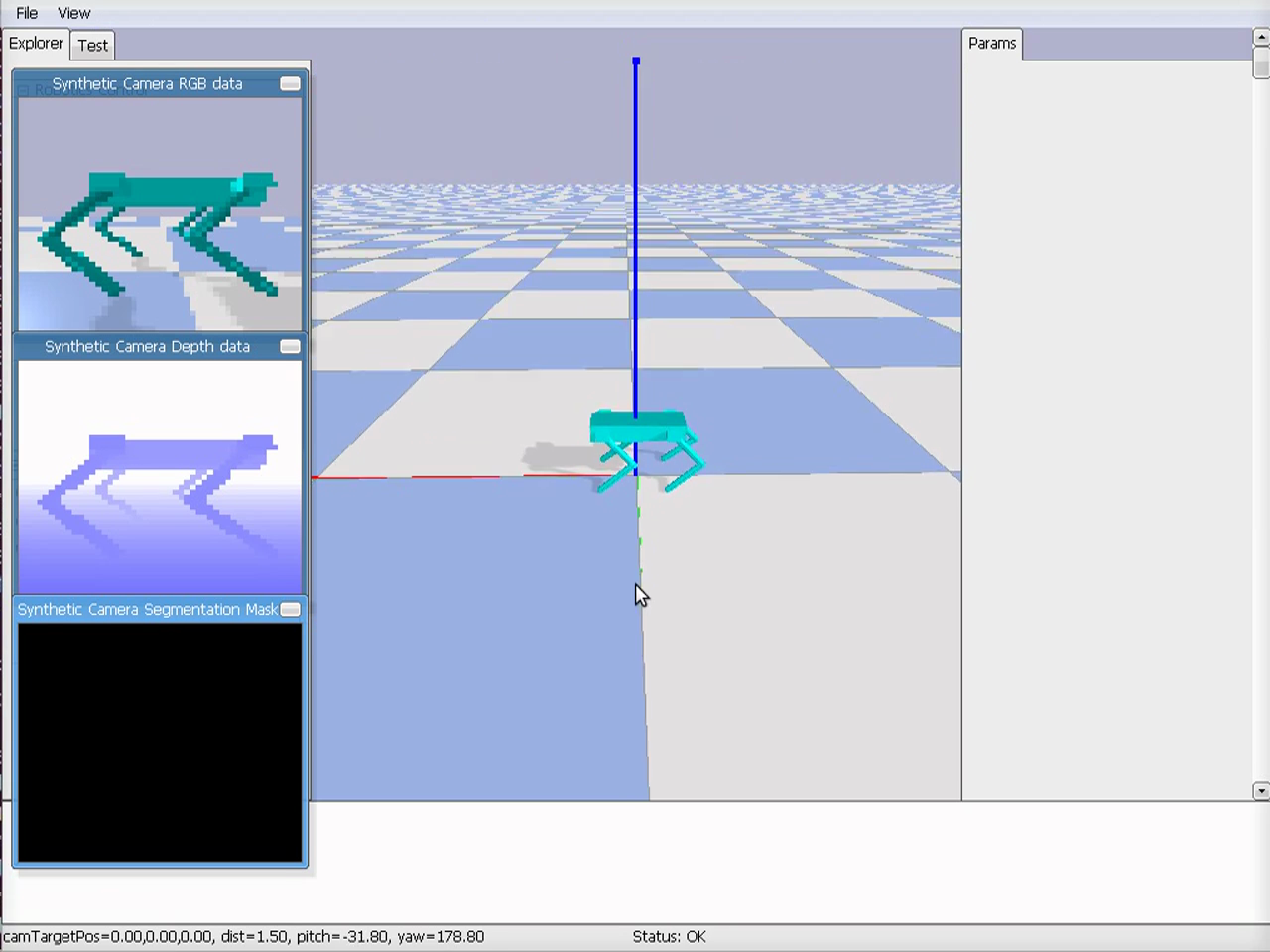}
\includegraphics[trim={11.0cm 8.0cm 11.0cm 13.0cm},clip,width=0.25\linewidth]{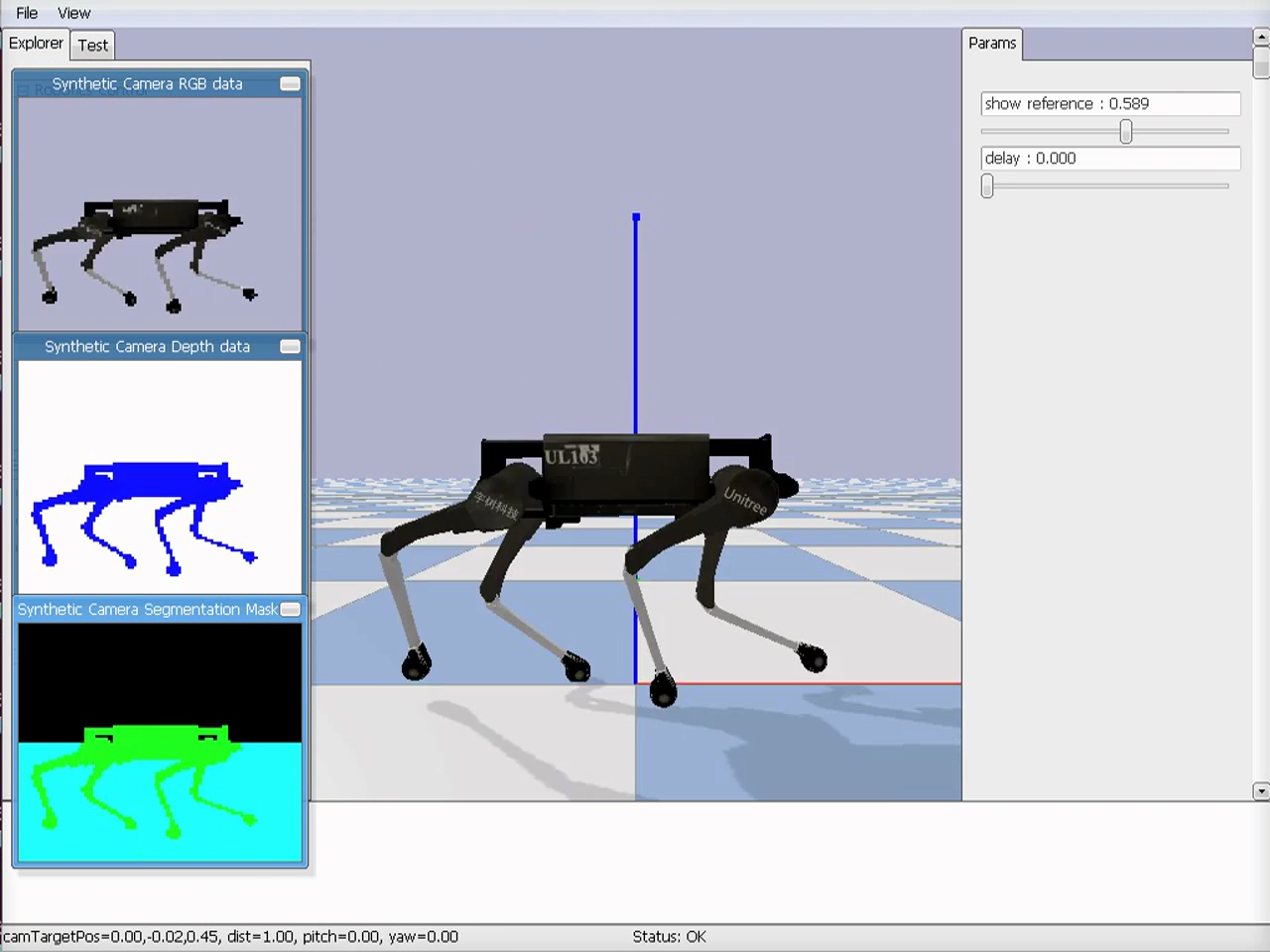}
\caption{ Pupper and Laikago agents, both of which are difficult platforms to learn policies on since they tumble easily and expert demonstrations are hard to obtain. 
}\label{fig:imitation_pupper} 
\end{SCfigure}
\paragraph{Sim2Real for Quadruped Robots}
\label{sec:sim2real}
We use \methodName to train policies for two simulated quadrupeds shown in~\refFigure{fig:imitation_pupper}. \changes{These environments have been used for Sim2Real transfer\citep{tan2018sim,RoboImitationPeng20}.} The resulting behaviours \changes{learned in simulation} are available at:~\href{\videoLink}{\videoLink}. We find that the Laikago environment is particularly challenging to learn; however, we can learn good policies on the smaller Stanford pupper in a day. \changes{This shows that \methodName can potentially be used to learn a control policy from a single video clip and transfer that policy to real hardware, however, we leave the details of the transfer process to future work.}

\section{Discussion and Conclusion}
\label{sec:discussion}

In this work, we have created a new method for learning imitative policies from a single demonstration.
The method uses a Siamese recurrent network to learn a distance function in both space and time.
This distance function learns to imitate video data where the agent's observed state can be noisy and partially observed. We use this model to provide rewards for training an \ac{RL} policy.
By using data from other motion styles and regularization terms, \methodName produces policies that demonstrate similar behaviour to the \demonstrationText in complex 3D control tasks. We found that the recurrent distance learned by \methodName was particularly beneficial when imitating demonstrations with greater partial observability.

One might expect that the \distanceMetricText should be pretrained to quickly understand the difference between a good and bad \demonstrationText. 
However, we have found that in this setting, learning too quickly can destabilize learning, as rewards can change, which can cause the \agent to diverge off to an unrecoverable policy space. 
In this setting, slower is better, as the \distanceMetricText may not yet be accurate. However, the learned distance function may be locally or relatively reasonable, which is enough to learn an acceptable policy. 
As learning continues, these two optimizations can converge together.

When comparing our method to \ac{GAIfO}, we have found that \ac{GAIfO} has limited temporal consistency. 
\ac{GAIfO} led to learning jerky and overactive policies.
The use of a recurrent discriminator for \ac{GAIfO}, similar to our use of sequence-based distance, may mitigate some of these issues and is left for future work.
It is challenging to produce results better than the carefully manually crafted reward functions used by the \ac{RL} simulation environments that include motion phase information in the observations~\citep{2018-TOG-deepMimic,Peng:2017:DDL:3072959.3073602}.
However, we have shown that our method can compute distances in space and time and learns faster than current methods that can be used in this domain.
A combination of beginning learning with our method and following with a manually crafted reward function could potentially lead to faster learning of high-quality policies if true state information is available.
Still, as environments become increasingly more complex and real-world training becomes an efficient option, methods that can learn to imitate from only video demonstration enable access to a vast collection of motion information, extensive studies on multi-task learning, and give way to more natural methods to provide instructions to agents.

\paragraph{Acknowledgements and Disclosure of Funding}
We want to acknowledge funding support from NSERC, CIFAR, and ElementAI/Service Now and compute support from ComputeCanada and ElementAI/Service Now.

\bibliographystyle{natbib}
\bibliography{paper}
\newpage
\addcontentsline{toc}{chapter}{Appendix}
\clearpage
\newpage
\section{Appendix}
\label{sec:Appendix}

This section includes additional details related to \methodName.

\subsection{Imitation Learning}

Imitation learning is the process of training a new policy to reproduce the behaviour of some \expert policy.
\ac{BC} is a fundamental method for imitation learning.
Given an expert policy $\policySymbol_{E}$ possibly represented as a collection of trajectories $\trajectory <(\myState_{0}, \action_{0}), \ldots, (\myState_{T}, \action_{T}) >$ a new policy $\policySymbol$ can be learned to match these trajectory using supervised learning.
\begin{equation}
    \label{equation:BC}
    \max_{\theta} \expectation_{\policySymbol_{E}}[\sum_{\ttime = 0}^{T} \log \policySymbol(\action_{\ttime}| \myState_{\ttime}, \modelParametersPolicy )]
\end{equation}
While this simple method can work well, it often suffers from distribution mismatch issues leading to compounding errors as the learned policy deviates from the expert's behaviour during test time.        

\subsection{Inverse Reinforcement Learning}
\label{sec:IRL}
Similar to \ac{BC}, Inverse Reinforcement Learning (\ac{IRL}) also learns to replicate some desired, potentially \expert, behaviour. However, \ac{IRL} uses the \ac{RL} environment to learn a reward function that learns to tell the difference between the agent's behaviour and the example data.
Here we describe maximal entropy \ac{IRL}~\citep{Ziebart:2008:MEI:1620270.1620297}.
Given an expert trajectory $\trajectory <(\myState_{0}, \action_{0}), \ldots, (\myState_{T}, \action_{T}) >$ a policy $\policySymbol$ can be trained to produce similar trajectories by discovering a distance metric between the expert trajectory and trajectories produced by the policy $\policySymbol$.
\begin{equation}
    \label{equation-IRL}
    \max_{c \in C} \min_{\policySymbol} (\expectation_{\policySymbol}[c(\myState, \action)] - H(\policySymbol) ) - \expectation_{\policySymbol_{E}}[c(\myState, \action)]
\end{equation}
where $c$ is a learned cost function and $H(\policySymbol)$ is a causal entropy term.
$\policySymbol_{E}$ is the expert policy that is represented by a collection of trajectories.
\ac{IRL} is searching for a cost function $c$ that is low for the expert $\policySymbol_{E}$ and high for other policies.
Then, a policy can be optimized by maximizing the reward function $r_{\ttime} = -c(\myState_{\ttime}, \action_{\ttime})$.

\subsection{Data}
\label{sec:appendix-data}
We are using the mocap data from the “CMU Graphics Lab Motion Capture Database” from 2002 (http://mocap.cs.cmu.edu/). To be thorough, we provide the processing at length. This data has been preprocessed to map the mocap markers to a human skeleton. Each recording contains the positions and orientations of the different joints of a human skeleton and can therefore directly be used to animate a simulated humanoid mesh. This is a standard approach that has been widely used in prior literature \citep{gleicher1998retargetting,rosales2000learning,lee2002interactive}. To be precise: at each mocap frame, the joints of a humanoid mesh model are set to the positions and orientations of their respective values in the recording. If a full humanoid mesh is not available, it is possible to add capsule mesh primitives between each recorded joint. This 3D mesh model is then rendered to an image through a 3rd person camera that follows the center of mass of the mesh at a fixed distance.

For the humanoid experiments, imitation data for $24$ other tasks was used to help condition the \distanceMetricText learning process. 
These include motion clips for running, backflips, frontflips, dancing, punching, kicking and jumping along with the desired motion. The improvement due to these additional unsupervised training data generation mechanisms are shown in~\refFigure{fig:ablation-humanoid3d}. 

\changes{The Sim2Real environments do not include video demonstrations. To create video data we use a similar method as in the other simulations. The available motion capture data is used in the simulation to control a \textit{kinematic} character from which 3rd person video data of that agent is collected.}

\begin{figure*}[t!]
\centering

\adjustbox{trim={.4\width} {.20\height} {0.4\width} {.25\height},clip}{\includegraphics[width=0.39\linewidth]{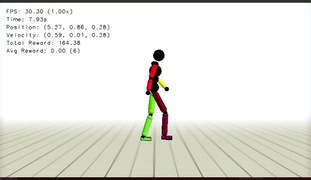}}
\adjustbox{trim={.4\width} {.20\height} {0.4\width} {.25\height},clip}{\includegraphics[width=0.39\linewidth]{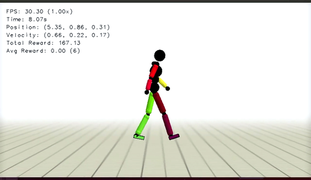}}
\adjustbox{trim={.4\width} {.20\height} {0.4\width} {.25\height},clip}{\includegraphics[width=0.39\linewidth]{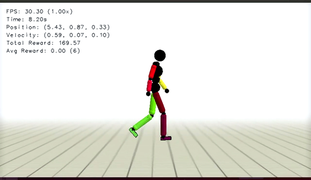}}
\adjustbox{trim={.4\width} {.20\height} {0.4\width} {.25\height},clip}{\includegraphics[width=0.39\linewidth]{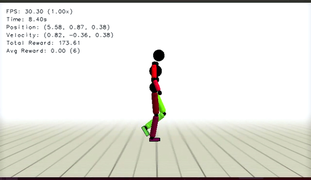} }
\adjustbox{trim={.4\width} {.20\height} {0.4\width} {.25\height},clip}{\includegraphics[width=0.39\linewidth]{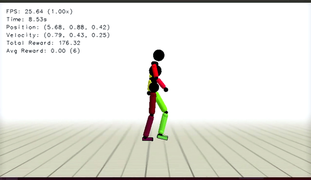} }
\adjustbox{trim={.4\width} {.20\height} {0.4\width} {.25\height},clip}{\includegraphics[width=0.39\linewidth]{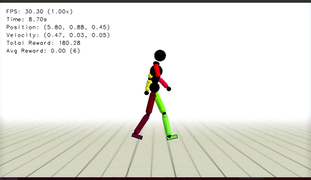} }
\adjustbox{trim={.4\width} {.20\height} {0.4\width} {.25\height},clip}{\includegraphics[width=0.39\linewidth]{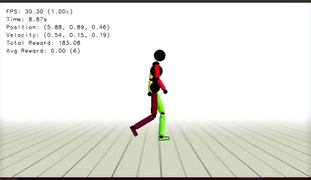} }
\adjustbox{trim={.4\width} {.20\height} {0.4\width} {.25\height},clip}{\includegraphics[width=0.39\linewidth]{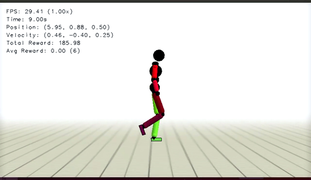} }
\adjustbox{trim={.4\width} {.20\height} {0.4\width} {.25\height},clip}{\includegraphics[width=0.39\linewidth]{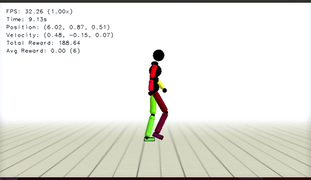} }
\adjustbox{trim={.4\width} {.20\height} {0.4\width} {.25\height},clip}{\includegraphics[width=0.39\linewidth]{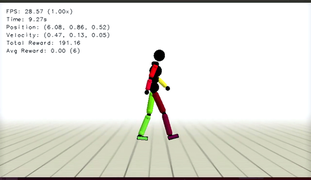} }
\adjustbox{trim={.4\width} {.20\height} {0.4\width} {.25\height},clip}{\includegraphics[width=0.39\linewidth]{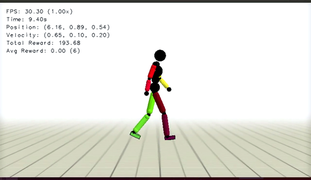} }\\

\adjustbox{trim={.4\width} {.20\height} {0.4\width} {.25\height},clip}{\includegraphics[width=0.39\linewidth]{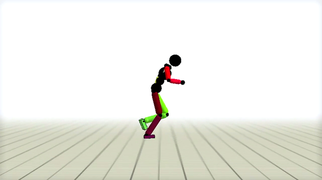} }
\adjustbox{trim={.4\width} {.20\height} {0.4\width} {.25\height},clip}{\includegraphics[width=0.39\linewidth]{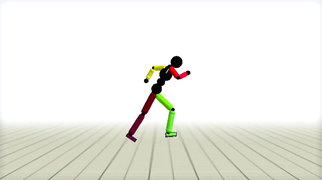} }
\adjustbox{trim={.4\width} {.20\height} {0.4\width} {.25\height},clip}{\includegraphics[width=0.39\linewidth]{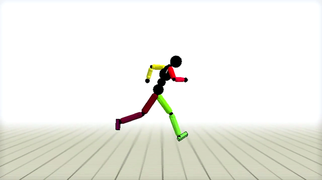} }
\adjustbox{trim={.4\width} {.20\height} {0.4\width} {.25\height},clip}{\includegraphics[width=0.39\linewidth]{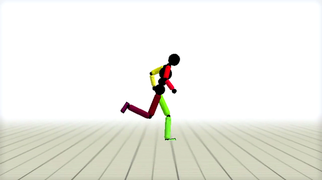} } 
\adjustbox{trim={.4\width} {.20\height} {0.4\width} {.25\height},clip}{\includegraphics[width=0.39\linewidth]{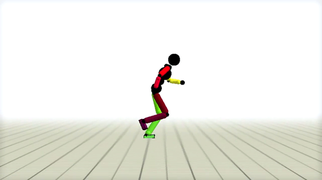} } 
\adjustbox{trim={.4\width} {.20\height} {0.4\width} {.25\height},clip}{\includegraphics[width=0.39\linewidth]{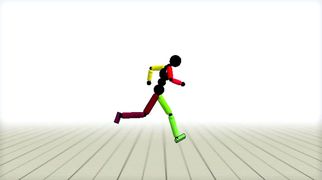} } 
\adjustbox{trim={.4\width} {.20\height} {0.4\width} {.25\height},clip}{\includegraphics[width=0.39\linewidth]{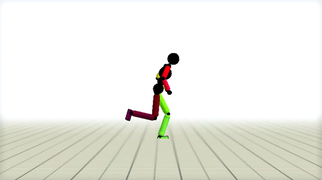} } 
\adjustbox{trim={.4\width} {.20\height} {0.4\width} {.25\height},clip}{\includegraphics[width=0.39\linewidth]{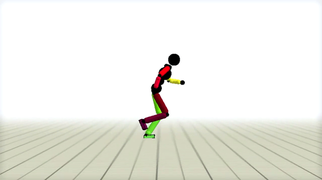} } 
\adjustbox{trim={.4\width} {.20\height} {0.4\width} {.25\height},clip}{\includegraphics[width=0.39\linewidth]{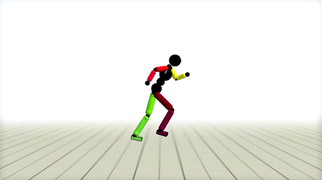} } 
\adjustbox{trim={.4\width} {.20\height} {0.4\width} {.25\height},clip}{\includegraphics[width=0.39\linewidth]{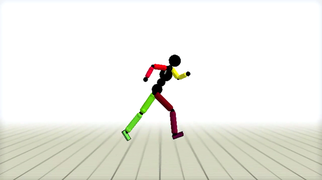} } 
\adjustbox{trim={.4\width} {.20\height} {0.4\width} {.25\height},clip}{\includegraphics[width=0.39\linewidth]{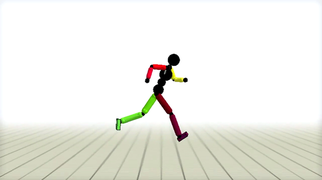} } \\

\adjustbox{trim={.4\width} {.20\height} {0.4\width} {.25\height},clip}{\includegraphics[width=0.39\linewidth]{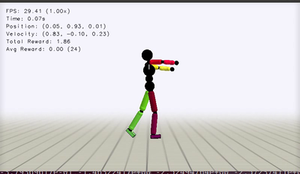}}
\adjustbox{trim={.4\width} {.20\height} {0.4\width} {.25\height},clip}{\includegraphics[width=0.39\linewidth]{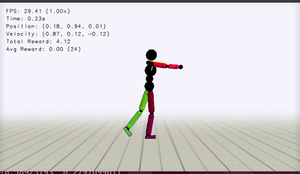}}
\adjustbox{trim={.4\width} {.20\height} {0.4\width} {.25\height},clip}{\includegraphics[width=0.39\linewidth]{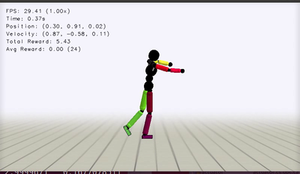}}
\adjustbox{trim={.4\width} {.20\height} {0.4\width} {.25\height},clip}{\includegraphics[width=0.39\linewidth]{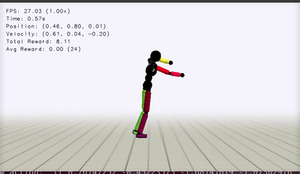}} 
\adjustbox{trim={.4\width} {.20\height} {0.4\width} {.25\height},clip}{\includegraphics[width=0.39\linewidth]{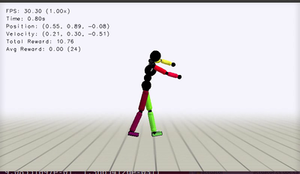}} 
\adjustbox{trim={.4\width} {.20\height} {0.4\width} {.25\height},clip}{\includegraphics[width=0.39\linewidth]{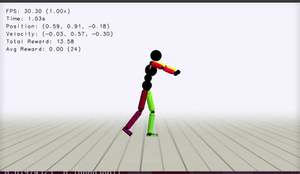}} 
\adjustbox{trim={.4\width} {.20\height} {0.4\width} {.25\height},clip}{\includegraphics[width=0.39\linewidth]{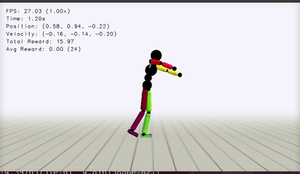}} 
\adjustbox{trim={.4\width} {.20\height} {0.4\width} {.25\height},clip}{\includegraphics[width=0.39\linewidth]{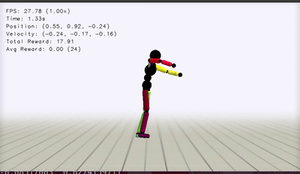} }
\adjustbox{trim={.4\width} {.20\height} {0.4\width} {.25\height},clip}{\includegraphics[width=0.39\linewidth]{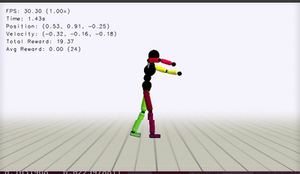}} 
\adjustbox{trim={.4\width} {.20\height} {0.4\width} {.25\height},clip}{\includegraphics[width=0.39\linewidth]{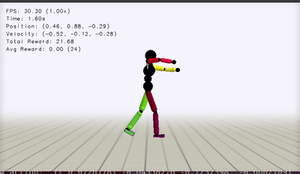}} 
\adjustbox{trim={.4\width} {.20\height} {0.4\width} {.25\height},clip}{\includegraphics[width=0.39\linewidth]{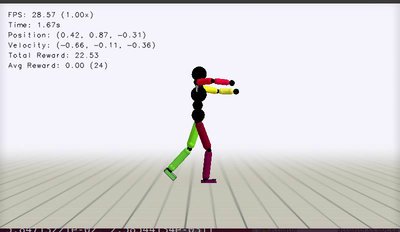}} \\

\adjustbox{trim={.4\width} {.20\height} {0.4\width} {.25\height},clip}{\includegraphics[width=0.39\linewidth]{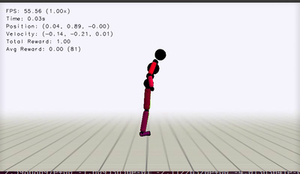}}
\adjustbox{trim={.4\width} {.20\height} {0.4\width} {.25\height},clip}{\includegraphics[width=0.39\linewidth]{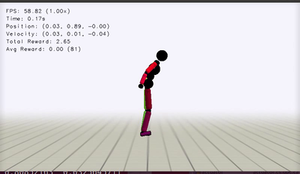}}
\adjustbox{trim={.4\width} {.20\height} {0.4\width} {.25\height},clip}{\includegraphics[width=0.39\linewidth]{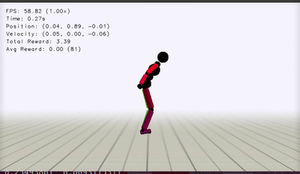}}
\adjustbox{trim={.4\width} {.20\height} {0.4\width} {.25\height},clip}{\includegraphics[width=0.39\linewidth]{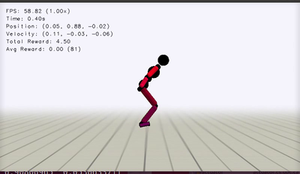}} 
\adjustbox{trim={.4\width} {.20\height} {0.4\width} {.25\height},clip}{\includegraphics[width=0.39\linewidth]{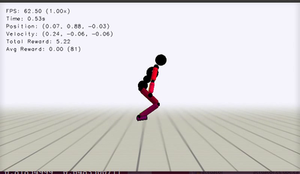}} 
\adjustbox{trim={.4\width} {.20\height} {0.4\width} {.25\height},clip}{\includegraphics[width=0.39\linewidth]{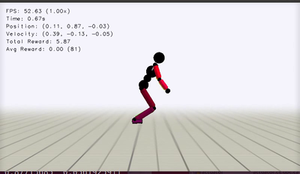}} 
\adjustbox{trim={.4\width} {.20\height} {0.4\width} {.25\height},clip}{\includegraphics[width=0.39\linewidth]{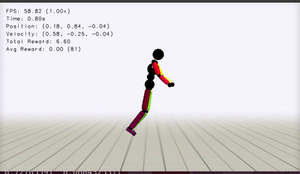}} 
\adjustbox{trim={.4\width} {.20\height} {0.4\width} {.25\height},clip}{\includegraphics[width=0.39\linewidth]{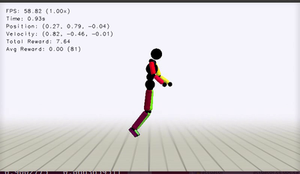} }
\adjustbox{trim={.4\width} {.20\height} {0.4\width} {.25\height},clip}{\includegraphics[width=0.39\linewidth]{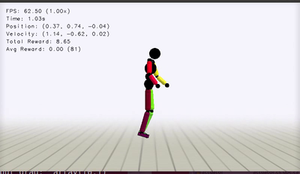}} 
\adjustbox{trim={.4\width} {.20\height} {0.4\width} {.25\height},clip}{\includegraphics[width=0.39\linewidth]{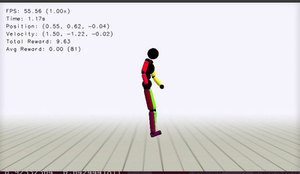}} 
\adjustbox{trim={.4\width} {.20\height} {0.4\width} {.25\height},clip}{\includegraphics[width=0.39\linewidth]{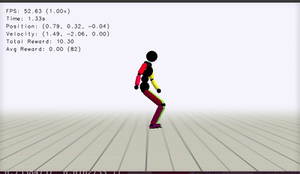}} \\

\caption{ \small
Rasterized frames of the imitation motions on \humanoidThreeD walking (row 1), running (row 2), zombie (row 3) and jumping(row 4).  ~\href{\videoLink}{\videoLink}
}
\label{fig:humanoid3d-motion}
\end{figure*}

\subsection{Training Details}

\begin{figure*}[t]
\centering 

\includegraphics[width=0.88\linewidth]{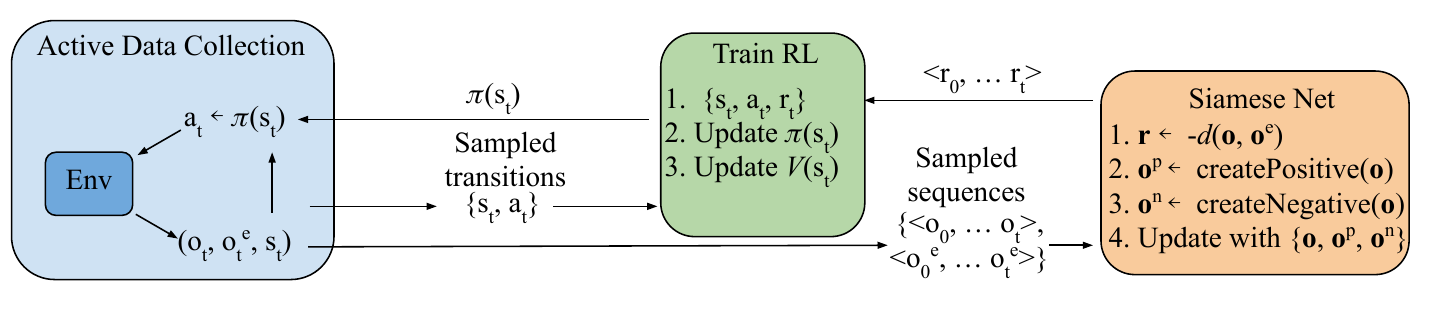}
\caption{\label{fig:virl-system} \small
The flow of control for the learning system.
}
\end{figure*}

The learning simulations are trained using \ac{GPU}s. The simulation is not only simulating the interaction physics of the world but also rendering the simulation scene to capture video observations. On average, it takes $3$ days to execute a single training simulation. 
The rendering process and copying the images from the GPU is one of the most expensive operations with \methodName.
We collect $2048$ samples between training rounds.
The batch size for \ac{TRPO} is $2048$.
The kl term is $0.5$.

The simulation environment includes several different tasks represented by a collection of motion capture clips to imitate.
These tasks come from the tasks created in DeepMimic~\citep{2018-TOG-deepMimic}.
We include all humanoid tasks in this dataset. \changes{The simulation uses RSI to randomly sample start states for the agent and expert to begin. This works by first uniformly randomly selecting a time in the expert demonstration and then synchronizing the learning agent with that time in the expert demonstration. This has shown to be very helpful across many prior papers to boost learning and is used for all algorithms in this paper.}

\begin{figure*}[!t]
\centering 

 \includegraphics[width=0.95\linewidth]{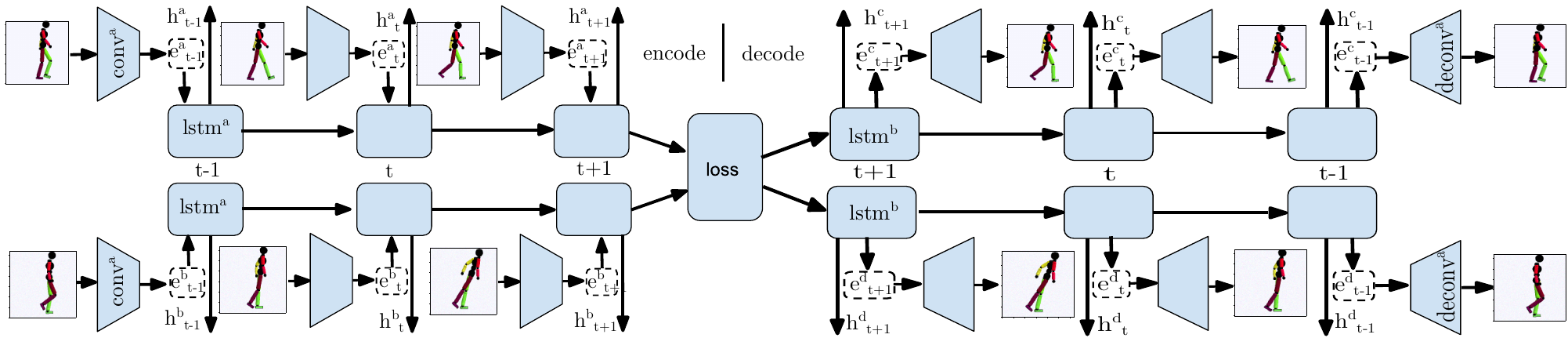} 
\caption{\small We use a Siamese autoencoding network structure that can provide a reward signal at every step to a reinforcement learning algorithm.  
For the Humanoid experiments here, the convolutional portion of our network includes $2$ convolution layers of $8$ filters with size $6 \times 6$ and stride $2 \times 2$, $16$ filters of size $4 \times 4$ and stride $2 \times 2$. The features are then flattened and followed by two dense layers of $256$ and $64$ units. 
The majority of the network uses \ac{ReLU} activations except for the last layer that uses a sigmoid activation. Dropout is used between convolutional layers.
The \ac{RNN}-based model uses a \ac{LSTM} layer with $128$ hidden units, followed by a dense layer of $64$ units.
The decoder model has the same structure in reverse, with deconvolution in place of convolutional layers.
}
\label{fig:siamese-models}
\vspace{-0.25cm}
\end{figure*}

In~\refAlgorithm{alg:VizImitation} we include an outline of the algorithm used for the method and a diagram in~\refFigure{fig:virl-system}. The simulation environment produces three types of observations, $\myState_{\ttime+1}$ the agent's proprioceptive pose, $\textbf{o}^{a}_{\ttime+1}$ the image observation of the agent and $\textbf{o}^{e}_{\ttime+1}$ the image-based observation of the expert demonstration. The images are grayscale $48\times48$. In \autoref{fig:siamese-models} we show a diagram of the network model using example data from the the \humanoidThreeD walking task. Different network structures were evaluated, this structure with the loss defined in \autoref{eq:virl} provided the best performance.

\subsection{Distance Function Training}

\begin{figure}[ht!]
\centering
\subcaptionbox{\label{fig:siamese-curve-image-based} Siamese loss during pretraining}{ \includegraphics[trim={0.0cm 0.0cm 0.0cm 0.0cm},clip,width=0.42\columnwidth]{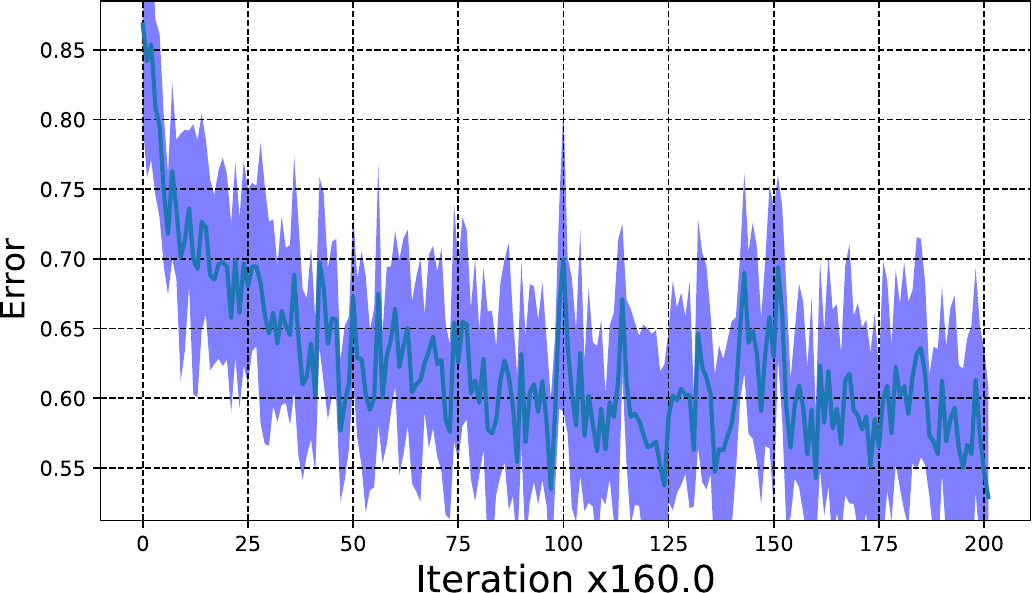}}
\subcaptionbox{\label{fig:siamese-curve-sequence-based} Siamese loss after pretraining}{ \includegraphics[trim={0.0cm 0.0cm 0.0cm 0.0cm},clip,width=0.42\columnwidth]{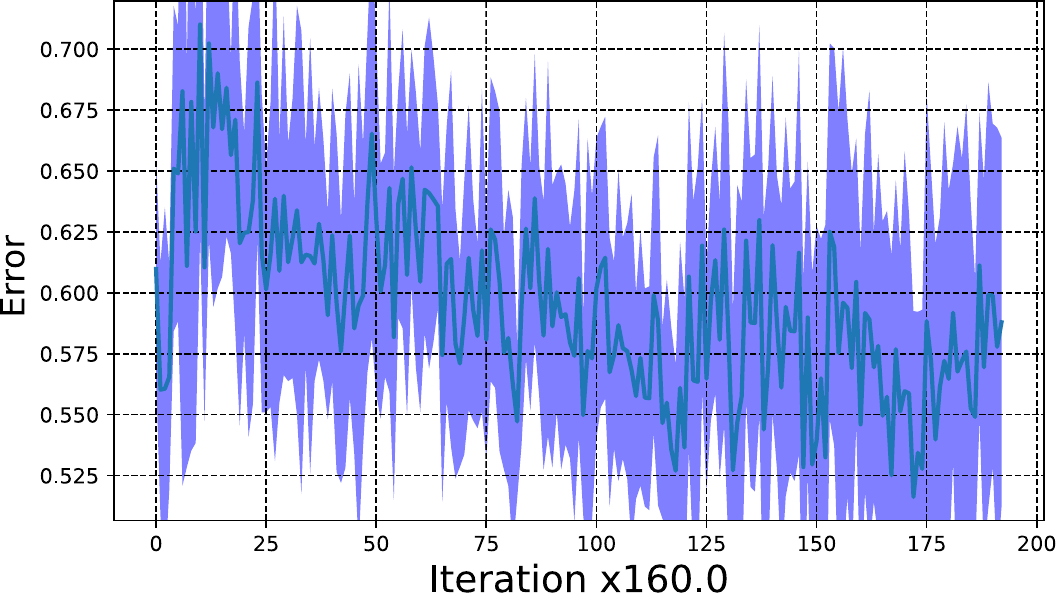}}
\caption{
Training losses for the Siamese distance metric.
}
\label{fig:siamese-curve}
\end{figure}

In~\refFigure{fig:siamese-curve-image-based}, the learning curve for the sequence-based Siamese network is shown during a pretraining phase.
We can see the overfitting portion the occurs during \ac{RL} training.
This overfitting can lead to poor reward prediction during the early phase of training.
In~\refFigure{fig:siamese-curve-sequence-based}, we show the training curve for the recurrent Siamese network after starting training during RL. After an initial distribution adaptation, the model learns smoothly, considering that the training data used is continually changing as the \ac{RL} \agent explores.

It can be challenging to train a sequenced-based distance function. 
One particular challenge is training the distance function to be accurate across the space of possible states.
We found that a good strategy was to focus on the earlier parts of the episode. 
When the model is not accurate on states earlier in the episode, it may never learn how to get into good states later, even if the distance function understands those better.  
Therefore, when constructing batches to train the \ac{RNN} on, we give a higher probability of starting earlier in episodes \ac{EESP}.
We also give a higher probability of shorter sequences as a function of the average episode length.
As the \agent gets better average episodes length increases, so to will the randomly selected sequence windows.

\changes{We found in our experiments that keeping the same \textit{in-order} sequence for decoding forced the encoding model to encode long-term temporal information from the beginning of training. This is particularly challenging to cope with as the RL policy is exploring different state distributions, further exasperating the challenging problem of learning good temporal representations. Instead, we reverse the decoding sequence which allows the training model to pickup on shorter term temporal dependencies quickly. This shorter but \textbf{more consistent} representation provides better signal to the RL agent. 
}
\subsection{Distance Function Use}

\begin{figure*}[ht]
\centering
\subcaptionbox{\label{fig:humanoid2d-reward-smoothing} Reward smoothing}{ \includegraphics[trim={0.0cm 0.0cm 0.0cm 0.0cm},clip,width=0.40\linewidth]{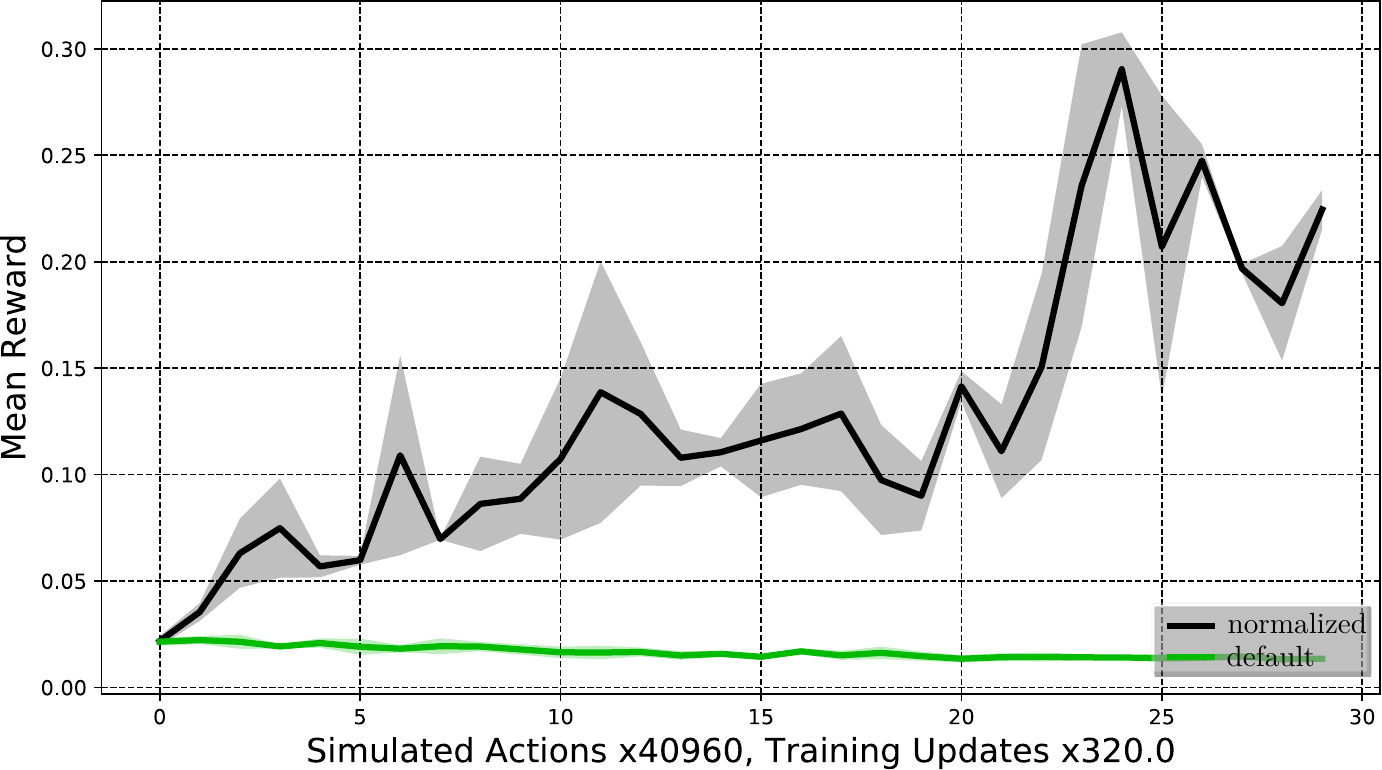}}
\subcaptionbox{\label{fig:baselines-cacla} Comparisons with other reward methods on \humanoidTwoD}{ \includegraphics[trim={0.0cm 0.0cm 0.0cm 0.0cm},clip,width=0.40\linewidth]{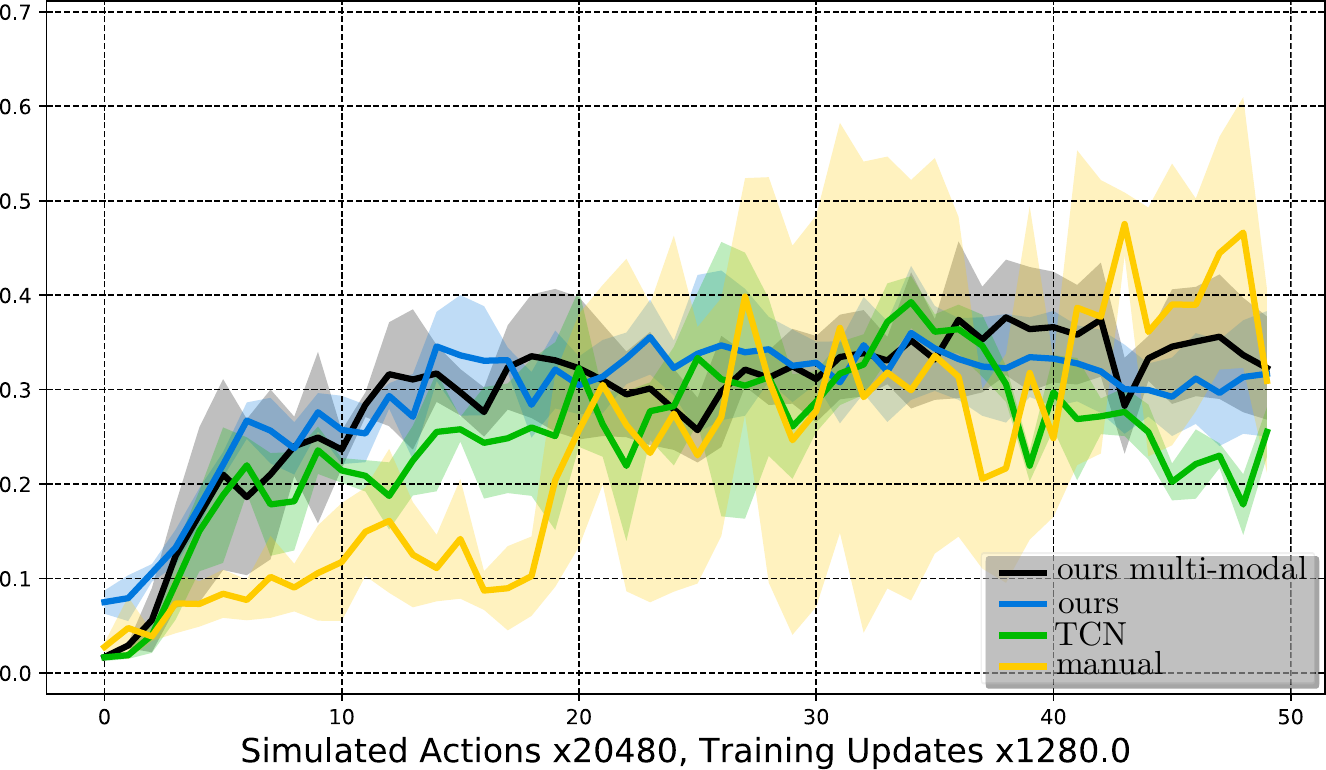}}
\caption{
Ablation analysis of \methodName.
We find that training \ac{RL} policies is sensitive to the size and distribution of rewards.
We compare \methodName to several baselines.
}
\label{fig:humanoid2d-hyperparam-analysis}
\end{figure*}

We find it helpful to \textit{normalize} the \distanceMetricText outputs using $\reward = \exp(\reward^{2} * w_{d})$ where $w_{d} = -5.0$ scales the filtering width.
This normalization is a standard method to convert distance-based rewards to be positive, which makes it easier to handle episodes that terminate early~\citep{Peng:2017:LLS:3099564.3099567,2018-TOG-deepMimic,peng2018variational}.
Early in training, the \distanceMetricText often produces large, noisy values. The \ac{RL} method regularly tracks reward scaling statistics; the initial high variance data reduces the significance of better \distanceMetricText values produced later on by scaling them to small numbers.
The improvement of using this normalized reward is shown in~\refFigure{fig:humanoid2d-reward-smoothing}. In~\refFigure{fig:baselines-cacla} we compare to a few baseline methods. The \textit{manual} version uses a carefully engineered reward function from~\citep{Peng:2017:DDL:3072959.3073602}.

\subsection{Positive and Negative Examples}
\label{sec:appendix-pos-neg-examples}
We use two methods to generate positive and negative examples.
The first method is similar to \ac{TCN}, where we can assume that sequences that overlap more in time are more similar.
We generate two sequences for each episode, one for the \agent and one for the imitation motion.
Here we list the methods used to alter sequences for positive pairs.
\begin{enumerate}
\item Adding Gaussian noise to each state in the sequence (mean $= 0$ and variance $= 0.02$)
\item Out of sync versions where the first state from the first and the last ones from the second sequence are removed.
\item Duplicating the first state in either sequence
\item Duplicating the last state in either sequence
\end{enumerate}
We alter sequences for negative pairs by
\begin{enumerate}
\item Reversing the ordering of the second sequence in the pair.
\item Randomly picking a state out of the second sequence and replicating it to be as long as the first sequence.
\item Randomly shuffling one sequence.
\item Randomly shuffling both sequences.
\changes{\item Using one sequence from the expert and one from the agent. We call these adversarial sequences pairs.
\item In the examples that include additional motion classes, the negatives are selected from the other classes.}
\end{enumerate}

The second method we use to create positive and negative examples is by including data for additional classes of motion.
These classes denote different task types.
For the \humanoidThreeD environment, we generate data for walking-dynamic-speed, running, backflipping and front-flipping.
Pairs from the same tasks are labelled as positive, and pairs from different classes are negative.

\paragraph{Viewpoint Invariance}
\label{sec:viewpoint-invariance}
\changes{Because we perform many types of data augmentations, clipping, warping, cropping, etc, the video data can be collected from viewpoints with different angles and distances, as long as most of the agent is captured by the video. The data augmentations are designed to help increase the diversity of data, but they also result in VIRL being able to compute distances from noisy data from different view locations.}

\subsection{Hyper Parameter Analysis}
\label{sec:app-hyperparam-analysis}
\changes{To determine the best values for $\lambda_{1:4} = \{ 0.7, 0.1, 0.1, 0.1\}$ in \autoref{eq:virl} we perform grid search over possible values in $0.1$ increments where $\sum_{i=0}^{4} \lambda_{i} = 1$. This evaluation is performed over all tasks using 3 seeds on each task in \autoref{subsection:sim-env} and the parameters that results in the best learning performance are selected. The final parameters are designed to be robust to new environments and may not require additional tuning.}

\end{document}